# Efficient learning of large sets of locally optimal classification rules

Van Quoc Phuong Huynh[1] · Johannes Fürnkranz[1] · Florian Beck[1]

**Abstract**
Conventional rule learning algorithms aim at finding a set of simple rules, where each rule covers as many examples as possible. In this paper, we argue that the rules found in this way may not be the optimal explanations for each of the examples they cover. Instead, we propose an efficient algorithm that aims at finding the best rule covering each training example in a greedy optimization consisting of one specialization and one generalization loop. These locally optimal rules are collected and then filtered for a final rule set, which is much larger than the sets learned by conventional rule learning algorithms. A new example is classified by selecting the best among the rules that cover this example. In our experiments on small to very large datasets, the approach's average classification accuracy is higher than that of state-of-the-art rule learning algorithms. Moreover, the algorithm is highly efficient and can inherently be processed in parallel without affecting the learned rule set and so the classification accuracy. We thus believe that it closes an important gap for large-scale classification rule induction.

## 1 Introduction

In the wake of the success of machine learning algorithms that learn inscrutable black-box models—most notably deep neural networks—explainable and interpretable models have gained again in importance. A popular line of work, pioneered by the algorithms

---

Machine LearningLIME (Ribeiro et al., 2016) and SHAP (Lundberg & Lee, 2017) is concerned with finding local white-box explanations that approximate the learned global black-box model in a given neighborhood of an example. LORE (Guidotti et al., 2018) is such an approach, which has a particular focus on rule-based explanations.

However, there is also considerable criticism towards this approach. Most notably, Rudin (2019) argues that instead of devoting more efforts to explaining black-box models, it might be preferable to focus on improving algorithms that learn white-box models in the first place, most notably rule learning algorithms. Consequently, several new rule learning algorithms have recently emerged, which we will briefly review in Sect. 2.4. In fact, as the extracted local white-box models are typically not perfectly aligned with the underlying black-box models, not even in their defined local neighborhood, it is not entirely clear that such approximations can serve as a valid explanation. Algorithms such as GLOCALX (Setzu et al., 2021) or TREEEXPLAINER (Lundberg et al., 2019) aim at constructing global white-box models out of local explanations, but these models are typically less accurate than their underlying black-box models, and the question of whether they outperform white-box models that have been directly learned from data is, in our opinion, still open.

In this paper, we argue that one of the key insights that result from local explanation algorithms like LORE is that each example gets its own individual explanation. This property has, so far, hardly been exploited in predictive rule learning, which typically try to find a concise set of rules that explain the examples with as few rules as possible. A notable exception is the HARMONY algorithm (Wang & Karypis, 2006), which takes an instance-centric view in that it aims at finding the best classification rule for each example. Moreover, historically, the first rule learning algorithms were example-based: AQ (Michalski, 1969, 1973) selected a random example and found the best rule that covers this example. However, mostly for computational reasons, this was not repeated for every possible example, but all examples that could be explained by the same rule were removed before the selection of the next example. CN2 (Clark & Niblett, 1989) was among the first algorithms that explicitly changed this strategy, from finding the best rule for a given example to finding a rule that explains as many examples as possible. Note that for any given rule evaluation measure, the rule found by such a strategy is not necessarily the best rule for any of the examples that it covers.

Motivated by these observations, we propose a rule learning algorithm, named LORD (Locally Optimal Rules Discoverer), that attempts to find the best rule for every training example. It does that very efficiently, using N-lists (Deng & Lv, 2015), a state-of-the-art data structure for frequent itemsets mining. From this data structure, for each training example, the best rule covering this example is extracted using greedy search. The found locally optimal rules are collected and filtered for a rule-based classifier, and new examples are classified by selecting the best of the covering rules in the classifier. While the found rule may still be a local optimum for covering an example, we nevertheless claim that the objective of finding the best rule for each example already makes a difference in comparison to the common goal of finding a simple rule that is good for as many examples as possible, even if the found rule is maybe not the globally best. Our experiments demonstrate that the algorithm compares well to state-of-the-art rule learning algorithms, not only in terms of accuracy, but also with respect to efficiency, which we demonstrate by evaluating it on very large datasets.

The remainder of the paper is organized as follows. Section 2 provides a brief foundations and survey on recent work of rule learning. Section 3 presents our approach which is then followed up with experiments and discussions in Sect. 4 and conclusion in Sect. 5.

🖄 SpringerContent courtesy of Springer Nature, terms of use apply. Rights reserved.

## 2 Locally optimal rules

In this section, we will briefly recall the foundations of rule learning and our notational conventions (Sect. 2.1), recall the basic principles of classic rule learning algorithms such as AQ and CN2 (Sect. 2.2), which serve as the basis for the key idea behind our approach, which will be introduced in Sect. 2.3. This section concludes with a brief review of relevant work in inductive rule learning (Sect. 2.4), before we turn our attention to our efficient implementation of the LORD algorithm in the next section.

### 2.1 Problem definition and notational conventions

The problem of rule learning assumes a number of labeled training examples $E = \{e_1, \ldots, e_n\} = \{\langle \mathbf{x}_1, y_1 \rangle, \ldots, \langle \mathbf{x}_n, y_n, \rangle\}$. Each training example $e = \langle \mathbf{x}, y \rangle$ consists of an instance $\mathbf{x}_i$ with its corresponding label $y_i \in C$, where $C$ is a nominal class attribute. The instances $\mathbf{x}_i$ are characterized with a set $F$ of $k$ binary *features*, i.e., $\mathbf{x}_i = \langle f_{i,1}, \ldots, f_{i,k} \rangle \in \{0,1\}^k$.

A *rule* $r$ maps a subset $F_r \subset F$ of the features to a label $y \in C$, i.e., it has the form

$$r : F_r \to y \tag{1}$$

with the semantics that every example $\mathbf{x}$, for which the features in $F_r \subset F$ (the *body* of the rule) are present (i.e., have the value 1), should be assigned the label $y$.[1]

**Definition 1** (Rule Coverage) A rule $r$ is said to *cover* an example $\mathbf{x}$ iff $F_r \subset F_\mathbf{x}$, where $F_r \subset F$ are the conditions in the body of the rule, and $F_\mathbf{x} \subset F$ are the features of example $\mathbf{x}$. We denote the set of examples in $E$ that are covered by the rule $r$ with $E_r \subset E$.

The learning problem consists of finding a set of rules $R$, which can be used to assign the correct labels to new, unseen data. Examples assigned to the correct class are called *true positives* and those assigned to a different class *false positives*.

In the following, we will assume tabular datasets that are based on a set of $m$ categorical attributes $A_i$, $i \in \{1, \ldots, m\}$, each of which has a fixed set of possible values $a_{ij}$, $j \in \{1, \ldots, |A_i|\}$. Note that this assumption is not crucial for the key idea of locally optimal rule induction, but it will be exploited by the efficient implementation introduced in Sect. 3. The set of features is then defined via all possible selectors.

**Definition 2** (Selector) A selector $s_{ij}$ is a single condition represented by $A_i = a_{ij}$ which selects examples (data rows) having value $a_{ij}$ for attribute $A_i$ from the input dataset.

The term *selector* goes back to Michalski (1973, 1983), who used it as a generalization of any type of comparison of an attribute value with a constant. It can be thought of as a binary feature, similar to the notion of an *item* in frequent pattern mining, but maintaining the association to its defining attribute $A_i$. In the following, we will nevertheless often use these two terms interchangeably.

---

[1] Without loss of generality, we assume that features cannot be negated. If necessary, we can model negation by including for each possible feature also its negation into the feature set $F$.

For simplicity, we simply ignore missing values, i.e., no selector will cover such a value, but other, more elaborate techniques are possible (Wohlrab & Fürnkranz, 2011). Numerical attributes can be handled by discretization (García et al., 2013).

### 2.2 Learning rule sets

Covering (aka *separate-and-conquer*) rule learning algorithms learn one rule at a time. For doing so, it searches for a rule that optimizes some quality criterion. When a new rule is found, all covered examples which satisfy all conditions in the body of the rule are removed from the training set, and the rule learning continues with another rule until all training examples are covered or the given stopping criteria are met. The difference among algorithms in this family is mainly in the way a new rule is found. This crucially depends on the search strategy (e.g., hill-climbing, beam search, or exhaustive search), and the heuristic criterion $h(.)$ that is used for evaluating rules (Fürnkranz, 1999).

---

**Algorithm 1** AQ-type rule induction

1: **function** AQ($E,F,h$)
2:    $R = \emptyset$
3:    **while** $E \neq \emptyset$ **do**
4:       randomly select example $\langle \mathbf{x}, y \rangle \in E$
5:       $r \leftarrow \arg\max_{r'=(B \rightarrow y), B \subset F_{\mathbf{x}}} h(r')$
6:       $R \leftarrow R \cup \{r\}$
7:       $E \leftarrow E \setminus E_r$
8:    **end while**
9:    **return** $R$
10: **end function**

---

Historically, AQ (Michalski, 1969) can be considered as the ancestor of this family of algorithms. It proceeds by selecting a random example that is not yet covered by any of the previously found rules, and search the space of all generalizations of this rule using a top-down beam search for finding the best rule. An abstract pseudo-code of the general idea behind AQ is shown in Algorithm 1. Note that line 5 aims at finding the rule that optimizes the heuristic function $h(.)$ for a randomly selected seed example.[2] However, once such an optimal rule has been found for the example $\mathbf{x}$, all examples covered by this rule will also be classified using this rule, and are thus removed from the training set (line 7). Note that for any such example $\mathbf{x}' \neq \mathbf{x}$, a better rule than $r$ may exist, but will not be searched unless it also happens to cover $\mathbf{x}$. So, in summary, AQ strives for finding optimal rules for some of the given training examples, and uses those for classifying all other examples to which they apply, regardless of whether they are the optimal rule for these examples or not.

CN2 (Clark & Niblett, 1989) took this one step further by combining the covering loop of AQ with ideas from decision tree learning, which are optimized collectively over all examples. The resulting algorithm is sketched as Algorithm 2. Instead of optimizing rules for individual examples, the algorithm strives for finding the best overall rule for the current set of training examples. The crucial difference is that no seed examples are selected (line 4 of Algorithm 1), and the best rule is searched over all possible rules that can be formed from all possible features and all possible classes (compare line 4 of Algorithm 2 to line 5 of Algorithm 1). Again, the details of the algorithm differ (e.g., later many of its successors optimize rules for each individual class $y \in C$ instead of optimizing over all classes, and the search for the best rule is in many cases greedy), but the key idea is to find a generally good rule, as opposed to finding the optimal rule for each individual example.

---

[2] The fact that AQ uses greedy search to approximate this optimum is not relevant to our argument. The inductive logic programming system Progol (Muggleton, 1995), e.g., uses essentially the same idea in a first-order logic setting, but uses an efficient exhaustive best-first search to find the optimum.

**Algorithm 2** CN2-type rule induction

```
1: function CN2(E,F,h)
2:     R = ∅
3:     while E ≠ ∅ do
4:         r ← arg max_{r'=(B→y), B⊂F, y∈C} h(r')
5:         R ← R ∪ {r}
6:         E ← E \ E_r
7:     end while
8:     return R
9: end function
```

This strategy is essentially still in use by many state-of-the-art rule learning algorithms. Most notably, the well-known RIPPER rule learning algorithm (Cohen, 1995) follows this strategy with some important enhancements. In particular, it does not solely rely on the choice of a suitable heuristic for fighting overfitting, but employs additional pruning and optimization techniques. Inspired by *incremental reduced error pruning* (Fürnkranz and Widmer, 1994), RIPPER effectively deals with the over-fitting problem by simplifying rules on a separate pruning set, and with additional loops of post-processing for optimizing a rule set. Particularly, the key idea is to examine whether or not to replace one rule from the previously learned rule set by a revised one which is formed by a growth and then a pruning phase aiming at reducing error of the entire rule set. RIPPER can still be considered the state-of-the-art in inductive rule learning, and is hard to beat in both predictive accuracy and the simplicity of the learned rule sets.

In general, the drawback of this family of techniques is that later rules are found just based on gradually reduced parts of the training set, resulting in that rules are discovered on insufficient statistic information. Also, the inherent sequence in the rule search makes the techniques harder to tackle big datasets which usually require to be processed in parallel.

### 2.3 Locally optimal rule learning

As we have seen in the previous section, common rule learning algorithms strive for finding rules that classify many examples well, as opposed to finding rules that are optimal for each individual example. Locally optimal rule discovery, as proposed in this paper, aims at solving this problem by reverting back to the basic AQ algorithm. But instead of being satisfied with finding the best rule for some of the examples, we compose an ensemble of rules by finding the best rule for each training example. The resulting basic idea is sketched in Algorithm 3.

```
Algorithm 3 locally optimal rule induction
1: function LORD(E,F,h)
2:     R = ∅
3:     for all ⟨x, y⟩ ∈ E do
4:         r ← arg max_{r'=(B→y), B⊂F_x} h(r')
5:         R ← R ∪ {r}
6:     end for
7:     return R
8: end function
```

Note that there is no covering or removal of examples (as in line 7 of Algorithm 1 or line 6 of Algorithm 2), or selection of seed examples (as in line 4 of Algorithm 1). Instead, one optimal rule is learned for each individual example. The resulting rule set thus, in principle, consists of a set of rules, each being optimal locally for one of the training examples.

This basic intuitive idea has some obvious disadvantages. Most notably, it seems to be very inefficient to search for the best rule for each example. To that end, we will propose an efficient search technique, based on ideas from association rule discovery, which is able to greedily find a best rule for each example in very much the same way as conventional rule learners like CN2 or RIPPER, but, by exploiting efficient data structures, can strive for finding the optimum rule for every example instead of only a small subset of rules. Our experiments will demonstrate that the resulting algorithm is able to deal with very large datasets, which cause problems for state-of-the-art rule learning algorithms.

The other key problem is that the resulting rule set will be considerably larger than the rule sets found by conventional rule learning algorithms. We note in passing that the size of the rule set is not necessarily the same as the size of the example set, because the same rule may be optimal for multiple examples, and will consequently only be added once to the set. We will also introduce some mild pruning techniques for further reducing the size of the rule set. Nevertheless, the interpretability of the remaining, large rule sets is still problematic. However, we argue that this does not hold for the individual rules. Unlike, e.g., large rule sets that are derived from random forests, our rule sets consist of locally optimal rules, which are reminiscent of locally optimal explanations which have been recently proposed in explainable machine learning. Algorithms such as LIME (Ribeiro et al., 2016) or SHAP (Lundberg & Lee, 2017) strive to find local white-box explanations that approximate a learned global black-box model in a given neighborhood of an example. In particular, LORE (Guidotti et al., 2018) learns rule-based explanations from the data. While each of the found rules can be used for explaining a single example, collectively, these explanations cannot be considered as an interpretable, global explanation for the domain, and substantial efforts are required to extract an interpretable global rule-based model from local rule-based explanations (Setzu et al., 2021). While we do not explicitly address the issue of interpretability in this paper, our approach was motivated by these algorithms. The meaning of the entire rule set will be hard to grasp, but the individual rules may serve as explanations for the examples for which they are optimal.

### 2.4 Related work

In this section, we briefly review additional work in inductive rule learning and relate it to our approach. It is not necessary for understanding the key contribution of this paper, the quick reader may safely skip forward to Sect. 3, where we introduce our efficient implementation of the Lord algorithm.

Rule-based methods are among the popular technique classes in data mining and machine learning (Fürnkranz et al., 2012). The methods generally can be categorized into two types, descriptive and predictive rule learning that are based on the purpose of use of discovered rules. *Descriptive rule learning* aims at discovering patterns catching relations between a target variable and a set of explaining ones known as subgroup discovery (Klösgen, 1996; Atzmüller, 2015), or co-occurrences of sets of items with other sets of items known as association rule discovery (Agrawal et al., 1993; Hipp et al., June 2000). *Predictive rule learning* tries to generalize the training data into a collection of rules that can make predictions for new examples. While descriptive rule learning aims at statistical validity of the found rules, predictive rule learning focuses on predictive performance. Our method introduced here belongs to the category of predictive rule learning, which can be discriminated into two main families, *rule set construction*, where rules are incrementally added to a target theory, and *rule set selection*, where first a large set of rules is learned which is then filtered into a small set of classification rules.

### 2.4.1 Covering algorithms

Covering algorithms, aka *separate-and-conquer* learning, construct a rule set by learning one rule at a time, removing all covered examples, and repeating until all examples are covered. The AQ, CN2, and Ripper algorithms discussed above are their main proponents, but the family of algorithms is very large (Fürnkranz, 1999). We note that the covering approach may be viewed as a special case of additive learning algorithms such as boosting, where the weights of the examples are not restricted to be 1 (uncovered) or 0 (covered), but can take arbitrary values. This approach was pioneered by algorithms such as LRI (Weiss and Indurkhya, 2000) and Slipper (Cohen and Singer, 1999), the general framework of gradient boosting for rule learning was most clearly defined in Ender (Dembczyński et al., 2010). Recent additions to this family include Boomer (Rapp et al., 2020), which generalizes this approach to learning multi-label rules, and the algorithm of Boley et al. (2021), which replaced the greedy search for the best addition to the rule set with an efficient exhaustive search.

### 2.4.2 Associative classification

Most prominent among the rule set selection techniques is associative classification (see, e.g., (Liu et al., 1998, 2000; Li et al., 2001; Yin and Han, 2003)). These algorithms generally search for association rules with target classes in their head, which are then filtered by certain pruning conditions or heuristic measurements to form a predictive rule set for the classification. Note that this filtering is often essentially equivalent to a covering loop. For example, CBA sorts all rules according to a heuristic $h(.)$, and then selects one rule at a time until all examples are covered. A common disadvantage of this family of algorithms is that the classification performance depends strongly on the input parameter *support*. In principle, smaller supports provide higher classification accuracy, but often the number of patterns and so the time complexity and found rules for some datasets explode as well. The diversity in the number of frequent patterns among datasets makes it hard to select an appropriate support for the trade-off between the classification performance and calculation resources.

DDPMine (Cheng et al., 2008) uses a modified version of FP-Growth to search for a set of discriminative patterns, as measured by information gain. The algorithm applies a pruning method in which searching on a conditional database can be ignored if the upper bound information gain from the conditional database is not greater than the information gain of the currently best frequent itemset. All examples supporting the result itemset will be removed from the current FP-tree, and another iteration continues until the current FP-tree is empty. The resulting discriminative itemsets can then be used as input features for training a subsequent classifier, such as a support vector machine. It remains unclear whether they can also be used as a stand-alone rule set.

The Harmony algorithm (Wang & Karypis, 2006) is quite similar to our approach in that it shares the general idea of finding the best rule for each training example. However, while our work is based on ideas in classical classification rule learning, Harmony is firmly rooted in association rule discovery. At its core, Harmony exhaustively enumerates all possible rules that satisfy a given minimum support threshold, and checks for each rule whether it has a higher confidence than the current best rule for each of its covered examples. By introducing several effective pruning methods, Harmony has shown efficient execution and higher classification accuracy than the covering algorithm Foil (Quinlan, 1990) and the associative classification algorithm CPAR (Yin and Han, 2003). However, like all associative classification algorithms, the performance of Harmony crucially depends on the chosen support threshold, which essentially defines a trade-off between the optimization quality and the efficiency of algorithm: lower support values will lead to an exponential growth in the search space that is covered exhaustively whereas higher minimum support values may miss optimal rules that have a low coverage (the importance of low-coverage rule was first observed by Holte et al. (1989)). Thus, Harmony copes with the exponential size of the hypothesis space by reducing it with a suitable choice of the minimum support threshold and finding instance-based global optima in this reduced space, whereas our approach, Lord, deals with the complexity by replacing the exhaustive search for the global optimum with an efficient greedy search for a local optimum. Obviously, both approaches may miss the globally best rule, but in very different ways, and it will eventually depend on the domain which one is more effective. However, we argue that Lord is considerably more flexible, in that it, e.g., allows for easy parallelization, as the search of the optimum for each example is independent of all other examples, or facilitates the use of arbitrary rule learning heuristics whereas Harmony's pruning heuristics depend on the use of support and confidence, which are not particularly well suited for classification rule learning (Fürnkranz & Flach, 2005).

### 2.4.3 Modern rule learning algorithms

In the wake of the success of deep learning, some efforts also went into the design of efficient rule learning algorithms that optimize a given loss function. Many recent approaches (Dash et al., 2018; Su et al., 2016; Wang et al., 2017; Wang and Rudin, 2015; Yang et al., 2017; Letham et al., 2015) limit their work to binary classification rules, but instead of finding rules based on heuristic measures, they optimize the learned rule collection (set or list) according to an objective function with a search procedure aiming at finding a sparse rule set. Because the proposed optimization methods are computationally expensive, and the size of the search spaces explodes, they often operate on truncated search spaces in a greedy fashion. This loses the guarantee that the output rule collections are globally optimal. Also, the search for Boolean rules can take very long on large datasets, which limits their applicability to big data.

IDS (Lakkaraju et al., 2016) forms a rule set (a decision set in their terminology) by sampling on a candidate rule space corresponding to the cross product between the set of frequent itemsets and the set of classes. A candidate rule is selected based on optimizing a joint objective function that combines multiple criteria such as accuracy, rule count, rule length, coverage, etc. for the learned rule set. Like IDS, other methods are also related to associative classification, but use alternative techniques for searching for an optimized rule set (or list) from pre-mined rules. BRL (Letham et al., 2015) produces a Bayesian posterior distribution over permutations of a set of Bayesian association rules, each of which defines a prior distribution over classes for its rule head rather than a single class. From this, decision lists with high posterior probability are selected to classify new examples. Its successor, SBRL (Yang et al., 2017) further improves BRL's computational efficiency. Highly similar to BRL, FRL (Wang and Rudin, 2015) also learns a rule list. BRS (Wang et al., 2017) searches for an optimal rule set by first generating association rules from positive examples w.r.t. one out of two classes of the target binary attribute. To reduce the number of candidate rules, the association rules are filtered with some criteria such as applying an upper bound on rule length, ensuring that the false positive rate is smaller than the true positive rate, or maximizing information gain. A simulated annealing algorithm with a prior probability-based objective function is applied to search for an optimal rule set in the space of subsets from the candidate rules.

The approach by Su et al. (2016) applies integer programming to formulate a Hamming-distance-based objective function and performs block coordinate descent or linear programming relaxation to search for an optimized Boolean rule set. In a similar way, the CG rule learning algorithm (Dash et al., 2018) finds a Boolean rule set which minimizes the sum of the number of positive examples classified incorrectly and the number of clauses (rule bodies) in the space of all possible clauses covering negative examples. The complexity of the resulting rule sets, which is quantified as the sum of length of the rules, is bounded by a given input parameter to control the complexity and avoid over-fitting. Since this approach is only suitable for very small datasets, column generation (Barnhart et al., 1998) is used for tackling larger datasets, which allows to generate only a small subset of all possible rules explicitly.

All these recent techniques have in common that they strive for finding a minimal rule set, for the sake of interpretability. As argued at the end of the previous section, this is not our main objective: instead we aim for high predictive accuracy and scalability to large datasets. For this, we also embark on ideas from association rule learning, which are discussed in the next section.

## 3 The LORD algorithm

In this section, we describe LORD, our efficient implementation of locally optimal rule discovery, which allows us to find the best rule for each training example.

Like many other state-of-the-art algorithms (cf. Sect. 2.4), LORD draws upon some ideas from association rule learning. In particular, we make use of PPC-trees and N-lists, which can efficiently summarize counts of conjunctive expressions. N-lists are a data structure which was originally proposed to efficiently discover frequent itemsets from a dataset of transactions via the state-of-the-art algorithm PrePost$^+$ (Deng et al., 2012; Deng & Lv, 2015). We adapt the N-list structure to tabular datasets with attributes $A_i, i \in \{1, \ldots, m\}$ by using selectors as features, and to classification problems with the last attribute $A_m$ as the nominal class attribute. We also assume that the first $m-1$ predictive attributes $A_i, (i < m)$ are nominal, i.e., numeric attributes will be discretized beforehand, but can be missing (value NULL).

In the following, we describe our adaptation of PPC-trees (Sect. 3.2) and N-lists (Sect. 3.3) to a classification setting. In Sect. 3.4, we show how these structures can be used for implementing an efficient rule learning algorithm, which combines advanced ideas from algorithms such as CN2 or RIPPER to learn, prune, and optimize large rule sets, for which we present an efficient representation in Sect. 3.5. All introduced definitions and concepts are illustrated with an example in Sect. 3.6 and Fig. 1. Finally, the complexity of the algorithm is analyzed in Sect. 3.7.

### 3.1 Initialization

In a first pass, the dataset is scanned to count the frequency of all distinct selectors. Selectors from the predictive attributes (group 1) and those from the class attribute (group 2) are sorted locally in the ascending order of their frequencies. The sorted group 2 is then appended to the end of sorted group 1, yielding a global order $O$ of selectors. We use symbol $\prec$ to express the order relationship between two selectors, e.g. $s_1 \prec s_2$ means $s_1$ precedes $s_2$ in the order $O$. We assume that sets of selectors will always be ordered according to $O$.

**Definition 3** (Selector-set) A *selector-set* is a set of selectors in which selectors are in a predefined order $O$ and there are no two selectors from the same attribute. A *k-selector-set* $s_1 s_2 \cdots s_k$ is a selector-set having $k$ selectors in the order $O$, $s_1 \prec s_2 \prec \cdots \prec s_k$.

The selected order $O$ of selectors does not influence the correctness of calculating the support count of selector-sets, but it will affect the memory efficiency of the PPC-tree and N-lists (presented in the next sections). Because the support count of a rule is always calculated after the support count of the rule body, it is helpful to place selectors of group 1 before selectors of group 2 so that the support count calculations for the rule can utilize the previously computed results for the rule body (cf. also Definition 7 below).

### 3.2 PPC-trees

In a second pass, the dataset is scanned to construct the so-called *PPC-tree* structure, a prefix tree. The PPC-tree consists of so-called *PPC-nodes*, which contain a distinct selector that it associates, as well as frequency information which accumulates the number of examples (selector-sets) passing through the node when they are inserted into the tree.

**Definition 4** (PP-code and PPC-node) A *PPC-node* stores the following components:

- a *selector* to which the node is associated
- a *PP-code* ⟨*pre*, *post*⟩, which encodes in what place the node is encountered in a preorder or post-order traversal of the tree respectively
- a *frequency count* (*freq*), which encodes how many examples pass through this node in the tree

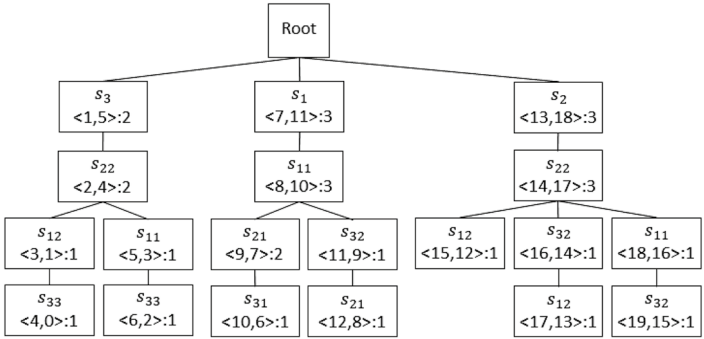

**Fig. 1** A running example from an example dataset to the result rule set

In addition, it also needs to store pointers to its parent and children nodes.

The PPC-tree is built up incrementally. Each example $e$ is represented with a sorted selector-set $S_e$. Note that it is possible that $|S_e| < m$ because $e$ can contain NULL values for some attributes, which are simply ignored. From the tree root, selectors in $S_e$ are inserted sequentially into the tree structure in the reverse order of $O$. When inserting a selector at a tree node, the child node registering the same selector is found and the frequency of the child node is increased by one. If there is no child node for the next selector, a new child node registering this selector is created with frequency initialized at 1. The inserting process continues with the next selector and the current node changed to the child node.

After all examples have been already inserted into the tree, the PP-codes are computed from a pre-order and post-order traversal of the tree. These allow to efficiently determine whether PPC-nodes are on the same path or not, using the following property of PP-codes (Deng and Lv 2015):

**Property 1** *Given two N-nodes $N_1$ and $N_2$ with their respective PP-codes $\langle pre_1, post_1 \rangle$ and $\langle pre_2, post_2 \rangle$, $N_2$ is an ancestor of $N_1$ (and thus $N_1$ is a descendant of $N_2$) iff $pre_2 < pre_1$ and $post_2 > post_1$.*

We will denote the sets of ancestors and descendants of a node $N$ with ANC($N$) and DESC($N$), respectively.

PPC-trees are similar to FP-trees (Han et al., 2004), which have also been adapted to supervised learning (Atzmüller and Puppe, 2006). Both are prefix trees constructed from a list of items ordered in increasing order of their global frequencies, with a similar way of inserting an example into the trees. Nodes of both trees contain an associated item and a frequency count. However, while each node of a PPC-tree contains a PP-code that allows to determine whether two nodes stay in the same path without looking at the tree structure, each node of an FP-tree contains a link to the next node of the same item, which allows to form a chain of nodes for each distinct item. Thus, an FP-tree has to maintain a header table to access the first nodes of node chains which supports to create conditional FP-trees. For this reason, FP-trees must also be retained throughout the entire process whereas the memory for PPC-trees can be freed after computing the more compact N-lists as described in the following section.

### 3.3 N-list generation

The PPC-tree is used for constructing the N-list of each selector, which effectively summarizes the frequency of occurrence of this selector. Thus, we associate with each selector a list of N-nodes, which collectively capture all matches of this selector in the data. After the generation of the N-list, the PPC-tree is no longer needed, and its memory can be freed.

**Definition 5** (N-node) An *N-node* is a reduced version of a PPC-node which only retains the PP-code and *freq*, denoted as $\langle pre, post \rangle : freq$.

**Definition 6** (N-list of a selector) The *N-list* of a selector is a list of all N-nodes associated with this selector. The N-nodes are sorted in increasing order of *pre*.

By a pre-order traversal in the built PPC-tree, an N-node is created from each visited tree node and added to the end of the N-list of the selector registered by the tree node. Completing the tree traversal, an N-list for each distinct selector is generated. No rearrangements to N-nodes in N-lists are performed since N-nodes were added in ascending order of *pre* while traversing the tree.

**Property 2** *Sorting the N-nodes in an N-list according to increasing pre, also sorts them according to increasing post.*

To see this, assume that the *pre* of two nodes in the N-list increases, but their *post* decreases. According to Property 1, the two nodes then have an ancestor-descendant relationship, i.e., they are on the same path in the PPC-tree. However, all nodes in an N-list share the same distinct selector, and therefore they all have to be on different paths of the PPC-tree, refuting the assumption.

The N-lists for single selectors, which can also be called 1-selector-sets, essentially correspond to the 1-itemsets in an Apriori-like association rule learning algorithm. However, as we see in the following, they contain all necessary information for constructing the N-lists for combinations of selectors.

For computing the N-list of a *k*-selector set, we use two different approaches:

**Definition 7** (N-list of *k*-selector-set, $k \geq 2$) The N-list $NL$ of the *k*-selector-set $s_1 s_2 \cdots s_{k-2} s_{k-1} s_k$ is calculated from the N-lists $NL_1$ of the $(k-1)$-selector-set $s_1 s_2 \cdots s_{k-2} s_{k-1}$, and $NL_2$ of either

(i) the $(k-1)$-selector-set $s_1 s_2 \cdots s_{k-2} s_k$, or
(ii) the selector $s_k$,

as follows

$$NL = \left\{ \langle N_2.pre, N_2.post \rangle : \sum_{\substack{N_1 \in NL_1 \cap \\ \text{DESC}(N_2)}} N_1.freq \ \middle| \ N_2 \in NL_2 \right\} \quad (2)$$

Thus, by this definition, the N-nodes in $NL$ sharing the same PP-codes will be combined into a single N-node with the same PP-codes and the sum of the frequency of the N-nodes.

The N-list calculation in Definition 7(i) provides sufficient information for an exhaustive search for frequent itemsets in which itemsets are enumerated in a unique order satisfying the input conditions for the calculation. However, in our greedy approach, we will need to be able to query and compute the support counts of arbitrary selector-sets, without the complete layer-wise enumeration of all itemsets that is typical for association rule discovery. For this reason, we use case (i) only if the N-list of $(k-1)$-selector-set $s_1 s_2 \cdots s_{k-2} s_k$ has already been computed from previous results. If it is not yet available, we use case (ii) the N-list of selector $s_k$ which is always available and better suited for greedy search.

In (2), a new N-node $N$ is added to the result N-list $NL$ for each pair of two nodes $N_1 \in NL_1$ and $N_2 \in NL_2$ that satisfies the ancestor-descendant relationship (cf. Property 1). $NL_2$ is always an ancestor of $NL_1$ because $N_1$ and $N_2$ associate with selectors $s_{k-1}$ and $s_k$ respectively and $s_{k-1} \prec s_k$ (cf. Definition 3). The new node $N$ receives the frequency count of $N_1$ (because it is the number of paths containing both $N_1$ and $N_2$) and the PP-code of $N_2$ (so that N-list of the selector-set reduces its length quickly thanks to node combinations at ancestor nodes while the selector-set grows, consequently reducing both memory consumption and computation time).

For implementing this, we define a recursive function *CalculateNList* (Algorithm 4), which can calculate the N-list of any selector-set $s_1 s_2 \cdots s_k$ as follows: *NListSet* is a hash map that caches generated N-lists and maps a selector-set to its N-list for fast access to N-lists of a given input selector-set. It initially contains all N-lists of every distinct selector (Definition 6). The function first checks if the N-list of the input selector-set has been calculated to return the N-list. Otherwise, it calculates recursively the N-list $NL_1$ of sub-selector-set $s_1 s_2 \cdots s_{k-2} s_{k-1}$ in lines 6-9. For lines 10-13, the N-list $NL_2$ will be the N-list of sub-selector-set $s_1 s_2 \cdots s_{k-2} s_k$ (Definition 7(i)) if it has been calculated and previously cached in *NListSet*; otherwise, $NL_2$ is assigned to the N-list of selector $s_k$ (Definition 7(ii)) which is usually longer than the N-list of selector-set $s_1 s_2 \cdots s_{k-2} s_k$ but avoids another branch of recursive calculation. At line 14, function *GenerateNlist* generates the N-list $NL$ from $NL_1$ and $NL_2$ based on Equation (2). All intermediate results are stored in *NListSet* so that the recursion stops as soon as a retrieved N-list is encountered.

For the correct and efficient calculation of the heuristic values for rule evaluations without the need of re-counting frequencies in the database, the following property is crucial:

**Property 3** *The support count of a k-selector-set constructed according to Definition 7 is the sum of frequencies of the N-nodes in its N-list.*

Note that this property is not trivial and does not hold for arbitrary ways of combining N-lists. Counter-examples and a proof for Definition 7(ii), which builds upon a previous proof of Definition 7(i) by Deng et al. (2012), Proposition 4, can be found in Appendix 1, so the property generally holds for LORD.

---

**Algorithm 4** Calculate N-list of a selector-set

```
 1: function CALCULATENLIST(s₁s₂ ··· sₖ, NListSet)
 2:     NL ← NListSet.GETNLIST(s₁s₂ ··· sₖ)
 3:     if NL ≠ ∅ then
 4:         return NL
 5:     end if
 6:     NL₁ ← NListSet.GETNLIST(s₁s₂ ··· sₖ₋₂sₖ₋₁)
 7:     if NL₁ = ∅ then
 8:         NL₁ ← CALCULATENLIST(s₁s₂ ··· sₖ₋₂sₖ₋₁, NListSet)
 9:     end if
10:     NL₂ ← NListSet.GETNLIST(s₁s₂ ··· sₖ₋₂sₖ)
11:     if NL₂ = ∅ then
12:         NL₂ ← NListSet.GETNLIST(sₖ)
13:     end if
14:     NL ← GENERATENLIST(NL₁, NL₂)
15:     NListSet.ADD(NL)
16:     return NL
17: end function
```

---

### 3.4 Rule learning

This subsection presents the proposed rule learning algorithm LORD (Locally Optimal Rules Discoverer). It first builds up the N-list structure for each distinct selector from the training set $E$ as discussed in the previous sections, which allows it to efficiently obtain the N-list and thus the coverage counts for an arbitrary rule body (Algorithm 4). This is used in a greedy search for a locally optimal rule, in a similar way as classical algorithms such as CN2 and RIPPER, as shown in Algorithm 5. In particular, the algorithm searches for a locally best rule for each training example in two phases, lines 6–13 for rule growth and lines 14–21 for rule pruning.

---

**Algorithm 5** LORD algorithm

**Input:** Training set $E$, heuristic $h(.)$
**Output:** Rule set $R$
1: Build N-list structure for each distinct selector from $E$
2: $R \leftarrow \emptyset$
3: **for** each example $e = \langle s_1 s_2 \cdots s_m, c \rangle \in E$ **do**
4:     $S_e \leftarrow \{s_1, s_2, \cdots, s_m\}, B \leftarrow \emptyset$
5:     $r : B \rightarrow c$
6:     **while** $S_e \neq \emptyset$ **do**     ▷ Rule Growth
7:         Find $s \in S_e$, so that $r' : B \cup \{s\} \rightarrow c$ is the best
8:         **if** $r' \succ r$ **then**
9:             $B \leftarrow B \cup \{s\}, S_e \leftarrow S_e \setminus \{s\}$
10:        **else**
11:            **break**
12:        **end if**
13:     **end while**
14:     **while** $|B| > 2$ **do**     ▷ Rule Pruning
15:         Find $s \in B$, so that $r' : B \setminus \{s\} \rightarrow c$ is the best
16:         **if** $r' \succ r$ **then**
17:             $B \leftarrow B \setminus \{s\}$
18:        **else**
19:            **break**
20:        **end if**
21:     **end while**
22:     $R \leftarrow R \cup \{r\}$
23: **end for**
24: $R' \leftarrow \emptyset$
25: **for** each example $e \in E$ **do**     ▷ Filter Rule Set
26:     $R_e \leftarrow \text{ClassCoveringRules}(R, e)$
27:     $R' \leftarrow R' \cup \{\text{BestRule}(R_e)\}$
28: **end for**
29: **return** $R' \cup \{\emptyset \rightarrow c_{majority}\}$

---

**Rule evaluation.** For comparing the quality of candidate rules, we use a heuristic function $h$, and coverage and class frequency for tie breaking.

**Definition 8** (Rule comparison) Given two rules $r_1$ and $r_2$, $r_1$ is better than $r_2$, denoted as $r_1 \succ r_2$ if (i) $h(r_1) > h(r_2)$ or (ii) $h(r_1) = h(r_2)$ and $r_1.p > r_2.p$ or (iii) $h(r_1) = h(r_2)$ and $r_1.p = r_2.p$ and $r_1.head \prec r_2.head$.

where $h(r)$ is the heuristic value, $r.p$ is the number of covered positive examples (true positives) of the rule $r$, and $r.head$ is the rule head of rule $r$. In the rare case that the heuristic value and the number of covered true positives are equal, we favor the rule with the minority class, because it covers a higher percentage of the positive examples of this class.

As a heuristic, we use the m-estimate, which has been proposed by Cestnik (1990) and used in CN2 (Clark and Boswell, 1991) and related algorithms (Džeroski et al., 1993). The m-estimate value of a rule $r : B \rightarrow c$ is calculated as

$$h_m(r) = \frac{r.p + m\frac{P}{P+N}}{r.p + r.n + m} \quad (3)$$

where

$m$ = a settable parameter in the range $[0, +\infty)$
$r.p$ = the number of true positives of rule $r$
$r.n$ = the number of false positives of rule $r$
$P$ = the number of positive examples in $E$ w.r.t. class $c$
$N$ = the number of negative examples in $E$ w.r.t. class $c$

The $m$-value is a tunable parameter that provides an excellent trade-off between weighted relative accuracy, which is frequently used in descriptive rule learning, and precision, the main target for predictive learning (Fürnkranz & Flach, 2005). It also has been shown to perform very well in a broad empirical comparison of various rule learning heuristics (Janssen & Fürnkranz, 2010). In LORD, we use comparably low values of $m$ (the default is 0.1), which result in a bias towards more specific and less general rules. This is analyzed in somewhat more depth in Sect. 4.7.

All counts can be efficiently obtained from the corresponding N-lists (Algorithm 4): For example, the number of true positives ($r.p$), or the total coverage of the rule ($r.p + r.n$) of rule $r$ can be computed efficiently from the support count of selector-sets $B \cup \{c\}$ and $B$ that are derived from their corresponding N-list structure according to Property 3. Similarly, $P$, the count of positive examples w.r.t class $c$, can be derived from the N-list of selector $c$.

The *CalculateNList* function defined above (Algorithm 4) is called frequently at lines 7 and 15 of Algorithm 5 to determine the support counts of selector-sets. In our implementation, the scope of the *NListSet* input of the function is limited to the search for the best rule of each example and not re-used across examples. However, these additional computational costs are compensated by the memory saved and the completely independent search for the best rule among examples, which allows for easy parallelization. This is empirically analyzed below, in Sect. 4.6.

**Rule growth**. In the rule growth phase, the rule body is first initialized with an empty set, and iteratively extended with the selector among the remaining selectors $S_e$ that results in the best improvement according to Definition 8. Note that, contrary to the similar search in algorithms like CN2 or RIPPER, the set of possible selectors does not contain all possible features, but only those that are pertinent for the current example, so that the complexity of this phase is bounded by the number of attributes $m$, and not by the considerably larger number of selectors that are derived from these attributes. The growth process finishes when the rule is no longer improved or $S_e$ is empty.

**Rule pruning**. In the pruning phase, selectors are iteratively pruned from the rule body so that the improvement in each step is maximized based on Definition 8. This is analogous to the incremental reduced error pruning technique (Fürnkranz and Widmer, 1994), which is used as in RIPPER, with the difference that all possible conditions are considered for pruning (not only final sequences), and the resulting rules are re-evaluated on the training set, not on a separate pruning set. The pruning phase completes when the rule cannot be further improved or when only two selectors remain in the rule body. In this case, we do not have to prune further, because all rules with a single condition have already been validated in the growing phase, and can therefore not have a higher heuristic value than the current rule.

As one of our astute reviewers has suggested, rules could be further improved by repeating multiple phases of rule specialization (growing) and generalization (pruning), or even a general bi-directional hill-climbing, which has, e.g., been tried in the JoJo algorithm (Fensel and Wiese, 1993). Empirically, however, this may not be necessary. The rules found in the growth phase often already achieve the highest heuristic value, so the chance to prune a selector from a rule in the pruning phase is small, and consequently, the chance to further improve the rule in a second growing phase (after the pruning phase) is even smaller. We will return to this issue in experiments reported in Sect. 4.8.

**Efficient Forward Rule Selection.** In order to reduce the computational complexity of LORD, we also implemented and tested a variant, LORD*, which only learns a rule when a training example is not correctly classified by the existing rule set. More precisely, LORD* does the following for each training example:

- If no rule in the current rule set covers the example, a new rule for the training example is learned and added to the current rule set.
- If the training example is mis-classified by the current rule set, a new rule is learned, and if the found best rule is better than the selected classifying rule, it is added to the current rule set.
- If the training example is correctly classified, no new rule is learned.

Note that a consequence of this is that the final rule set learned by LORD* is somewhat affected by the input order of training examples. However, under the assumption of a random order, it seems that the technique works well, as we will see in Sect. 4.

**Rule Filtering.** RIPPER introduced a very effective but rather expensive rule optimization phase. LORD also implements an optimization phase, in the form of a very simple filter that removes rules which have been superseded by better rules. All found best rules are collected in a rule set $R$ which is then filtered to the final rule set $R'$ in lines 24–28. CLASSCOVERINGRULES computes the set $R_e \subset R$ of all rules that cover $e$ and whose rule head equals the class label of $e$. The best rule from a group of rules is selected based on Definition 8. Selecting the best rule $r$ from $R_e$ further reduces the rule variances and therefore also the number of rules. Assume that there is a group of $k$ training examples sharing a global best rule. In the search phase, however, because of the greedy search, a different locally optimal rule may be found for each of the $k$ examples, e.g., a total of $l$ rules ($l \leq k$) with the best rule $r_0$. Then in the filter phrase, all the $k$ examples adopt $r_0$ as their best covering rule and the other $l - 1$ rules will be eliminated. Our experiments (Table 8) will show empirically that this hypothesis seems to hold and will underline the effectiveness of the rule filter. Generally, the larger the number of examples in the training set is, the higher is the chance that such a rule $r_0$ is actually the best rule. Finally, the default rule which has an empty body and predicts the majority class, is added to $R'$ to guarantee that every testing example is covered by at least one rule.

**Classification with the Learned Rule Set.** In the classification phase, the best rule is chosen from a group of rules covering the example based on Definition 8 to classify an unseen example. If no covering rule is found for the example, the default rule is used.

### 3.5 R-tree

For classifying a new example, LORD looks for the best rule among all rules that cover this example. In order to speed up this process, we index the result rule set via a so-called *R-tree*, a prefix tree of rule bodies whose selectors are in the same predefined order $O$ as the selector-sets (Definition 3). Each node (excluding the root node) of an R-tree associates with a distinct selector and may or may not contain a reference to a rule. Its child nodes are also in the order $O$. Selectors of a rule body are inserted into the tree in the reverse order of $O$. Thus, the tree serves as an index structure that allows to efficiently find the covering rules of an example whose corresponding selectors are also in the order $O$. The structure of an R-tree is quite similar to that of an FPO-tree (Huynh & Küng, 2020), which provides optimal compactness and efficient aggregations for very large numbers of local frequent itemsets. A minor difference is that the R-tree indexes references to rules whereas the FPO-tree records support counts of itemsets. Figure 1e shows an example of an R-tree.

### 3.6 Example

Figure 1 illustrates the entire process from an example dataset in Fig. 1a to the resulting rule set in Fig. 1e. The example dataset includes three predictive attributes $A_1$, $A_2$ and $A_3$, and the class attribute $C$. The right-most column in Fig. 1a depicts the selector-sets corresponding to examples. Figure 1b shows all distinct selectors in the predefined order $O$ (Sect. 3.1) after the first data scan. Figure 1c depicts the corresponding PPC-tree (Sect. 3.2) built from the example dataset shown in Fig. 1a.

The left column in Fig. 1d enumerates the N-lists (Sect. 3.3) of all single selectors that can be derived directly from the example PPC-tree in Fig. 1c. These basic N-lists are then combined to N-lists of selector-sets and their corresponding support counts. The middle and the right columns in Fig. 1d respectively show the generated N-lists and the corresponding candidate rules estimated while finding the best rule for the first example with its representative set of selectors $\{s_{31}, s_{21}, s_{11}\}$ and the class selector $s_1$. The best rule initializes with an empty body ($\emptyset \rightarrow s_1$) and is then extended with each of the three possible selectors $s_{31}, s_{21}, s_{11}$. For the selector $s_{31}$, with the candidate rule $s_{31} \rightarrow s_1$, the calculation is as follows:

1. Compute the N-list of selector-set $s_{31}s_1$ from that of $s_{31}$ and $s_1$ (Definition 7). N-node $\langle 10, 6 \rangle : 1$ is a successor of N-node $\langle 7, 11 \rangle : 3$ because it comes after $\langle 7, 11 \rangle : 3$ in the pre-order traversal ($10 > 7$) and before $\langle 7, 11 \rangle : 3$ in the post-order traversal ($6 < 11$). Consequently, a new N-node $\langle 7, 11 \rangle : 1$ is formed from the PP-codes of the ancestor and the frequency of the successor.
2. Compute the m-estimate (3) of the rule, yielding $h_1(s_{31} \rightarrow s_1) = 0.6875$ for $m = 1$.

The same calculation is applied for selectors $s_{21}$ and $s_{11}$, and eventually rule $s_{21} \rightarrow s_1$ is selected as the currently best rule because its heuristic value of 0.8437 is the highest. $s_{31}$ and $s_{11}$ are then considered as extensions. The evaluation of $h_1$ for the extended rules triggers the computation of the N-lists of $s_{31}s_{21}$ and $s_{31}s_{21}s_1$ (for the extended rule $s_{31}s_{21} \rightarrow s_1$) and of $s_{21}s_{11}$ and $s_{21}s_{11}s_1$ (for $s_{21}s_{11} \rightarrow s_1$). Their heuristic values, 0.6875 and 0.8437 respectively, do not exceed $h_1(s_{21} \rightarrow s_1)$. As a consequence, the rule growth stops. Also, the rule pruning is skipped in this case because the rule body length is 1.

In this way, the search for a locally best rule for each of the remaining examples is performed. All found rules are listed in the left two columns in Fig. 1e. Note that the first three examples share the same locally best rule $s_{21} \rightarrow s_1$. A filter step is then applied to the rule set in which rule $s_{12}s_{32} \rightarrow s_2$ of example #7 is removed because the example adopts rule $s_{32}s_{22} \rightarrow s_2$ of example #8 as its new best rule. The filtered rule set is then complemented with a default rule, which has an empty body and predicts the majority class $s_2$.

The right-most column in Fig. 1e illustrates an R-tree which has been constructed from the result rule set on its left. If none of these rules covers a new example, the default rule will be used to classify the example.

### 3.7 Complexity analysis

For the following analysis of LORD's computational complexity, we assume a dataset with $n$ examples and $m$ attributes. Furthermore, we use $k$ to denote the maximum length of a rule. Note that $k$ can be bounded in various ways. For example, obviously $k \leq m$, as each attribute is tested at most once. We can also assume that $O(k) \leq O(\log n)$, assuming that each condition reduces the number of examples covered by the rule by a certain fraction. Finally, $k$ can of course be bounded in practice by allowing only rules up to a certain fixed length $k$.

The number of rule evaluations in the growth phase is obviously $\sum_{j=0}^{k-1}(m-j) = m \cdot k - (k-1)k/2$ as the $j^{th}$ selector added to the rule body has $(m-j)$ options. Similarly, the pruning phase requires at most $\sum_{j=0}^{k-2}(k-j) = (k-1)k - (k-2)(k-1)/2 = (k-1)k/2 + 1$ rule estimations. Therefore, the total number of rule evaluations is $O(m \cdot k)$, which is considerably less than the $O(2^m)$, the number of all possible rules.[3]

A key factor for LORD's efficiency is that the rule evaluations do not have to be performed on data, but using the N-lists stored in memory. For computing the heuristic value of a candidate rule, $s_1 s_2 \ldots s_k \to c$, we need the N-lists of the two selector-sets $s_1 s_2 \ldots s_k$ and $s_1 s_2 \ldots s_k c$. First, the N-list of $s_1 s_2 \ldots s_k$ is calculated recursively by Algorithm 4 which, in the worst case, must calculate the N-lists of $k-1$ selector-sets $s_1 s_2, \cdots, s_1 s_2 \ldots s_{k-1}$, $s_1 s_2 \ldots s_k$ based on Definition 7. Since the PP-codes of the N-nodes in an N-list are in order, the calculation of the N-list of the $k$-selector-set in Definition 7 is linear in the length $l$ of the N-lists. Secondly, the N-list of $s_1 s_2 \ldots s_k c$ can be calculated directly from N-lists of $s_1 s_2 \ldots s_k$ and $c$ based on Definition 7(ii) (note that the N-list of $c$ has only one N-node). Overall, the computational complexity of a rule estimation for a rule $s_1 s_2 \ldots s_k \to c$ is $O(k \cdot l)$. Note that $l$ is bounded by the number of examples covered by the body of the rule and therefore depends on the density of the corresponding selectors in the dataset.

The overall computational complexity of LORD, therefore, is $O(n \cdot m \cdot k^2 \cdot i \cdot l)$. Note that the search for a local best rule for a training example is inherently independent of the other examples' searches. This allows the rule search to be parallelized massively with multi-workers, e.g. shared memory with multi-threads and/or distributed environment, in a work-pool model for high load balance. Thus, the total time complexity of LORD is $O(\frac{1}{q} \cdot n \cdot m \cdot k^2 \cdot i \cdot l)$, where $q$ is the number of parallel rule search workers.

---

[3] In Sect. 4.8, we will discuss OVERLORD, a variant which does multiple iterations of growing and pruning instead of a single one. This increases the complexity to $O(m \cdot k \cdot i)$, if $i$ iterations are performed. It is hard to find a theoretical bound for the number of iterations (in the worst case it could be $O(n)$). However, in our experiments, we found that on average only 5.4% of the training examples need more than one iteration of growing and pruning for finding a locally optimal rule in this setting, i.e., the average case of $i$ seems to be only slightly larger than 1.

## 4 Experiments

In this section, we report on the experimental evaluation of the LORD algorithm. The main goals of this study is to show that the algorithm is able to efficiently learn rule sets for very large databases, with an accuracy that is not worse and often better than that of other state-of-the-art algorithms (Sect. 4.2). We also compare the algorithms w.r.t. runtime (Sect. 4.3) and rule complexity (Sect. 4.4), and ensure that LORD's discretization does not give the algorithm an unfair advantage (Sect. 4.5). Furthermore, we do extensive experiments to analyze the scalability of LORD dealing with very large datasets (Sect. 4.6) and investigate the impact of the parameter $m$ of the m-estimate heuristic on the classification accuracy (Sect. 4.7), the potential of using multiple successive growing and pruning phases (Sect. 4.8), as well as the effect and potential of rule filtering on the rule sets learned by LORD (Sect. 4.9).

### 4.1 Experimental setup

Table 1 gives a brief characterization of the 24 UCI datasets (, 2017) used in the experiments. The first 12 datasets are small, and the remaining 12 range from medium to big volumes up to several millions of examples. The datasets are also quite diverse in terms of number and type of attributes, as well as their completeness and class distributions. The *gas-sensor-12* (Huerta et al., 2016) dataset contains data of gas sensors reacting to stimuli (wine, banana), and *gas-sensor-11* is a version of *gas-sensor-12* after removing the time-offset attribute for the duration between the start of the stimulus emission and the sensor operation that cannot be recorded as an input in the practice of gas detection. The dataset at row 23 is a cleaned version of *pamap2* (Reiss and Stricker, 2012) with 3,850,505 instances and 54 attributes, for physical activity monitoring, that was processed according to recommendations of the dataset's authors. The other datasets are used as they are.

In principle, any discretization method could be used to discretize numeric attributes, such as FUSINTER (Zighed et al., 1998), or the well-known MDLP (Fayyad and Irani, 1993). We selected the former because in our experiments, it ran faster than MDLP, and LORD can provide slightly higher prediction performance on discretized datasets by FUSINTER compared to that by MDLP. In Sect. 4.5, we double-check that the discretization did not give LORD an unfair advantage over the other algorithms.

We have compared LORD, encoded in Java, with a classification association rule learning algorithm CMAR (Li et al., 2001), the classic heuristic rule learning algorithm RIPPER (Cohen, 1995), and the two recent algorithms IDS (Lakkaraju et al., 2016) and CG (Dash et al., 2018). Despite its age, RIPPER is still among the state-of-the-art algorithms and is well known for its high accuracy, the simplicity of the learned rule sets, and its fast execution times. We use JRIP, an implementation of RIPPER, that can be found in the WEKA library.[4] IDS and CG are two rule learning algorithms proposed recently. For CG, we used the source code provided by the authors.[5] For IDS, we also started with the authors' source code,[6] but soon found that the IDS algorithm ran too slowly for our machine. It could not complete a cross-validation fold with the time-out setting, even on the small datasets. This is because the calculation for selecting a rule formed from frequent itemsets and classes makes IDS sensitive to the total number of frequent itemsets. A similar observation was also made by Filip and Kliegr (2019) who have customized IDS by allowing it to select only the top $k$ association rules instead of all. As their customized algorithm, PyIDS,[7] runs much faster, we have used it in our experiments, even though it cannot guarantee the same prediction performance as the original. For CMAR, we considered two implementations, respectively from the libraries SPMF[8] and LUCS-KDD[9] and report the results of the latter, which performed better in both runtime and accuracy.

---

[4] WEKA is available at https://waikato.github.io/weka-wiki/.
[5] https://github.com/Trusted-AI/AIX360.
[6] https://github.com/lvhimabindu/interpretable_decision_sets.

**Table 1** Datasets used in experiments

| # | Datasets | # Exs. | # Attr. | Attr. types | Missing values | Class distributions (%) |
|---|---|---|---|---|---|---|
| 1 | Lymph | 148 | 19 | Categorical | No | 54.7; 41.2; 2.7; 1.3 |
| 2 | Wine | 178 | 14 | Numeric | No | 33.2; 39.9; 26.9 |
| 3 | Vote | 435 | 17 | Categorical | Yes | 54.8; 45.2 |
| 4 | Breast-cancel | 699 | 10 | Numeric | Yes | 65.5; 34.5 |
| 5 | Tic-tac-toe | 958 | 10 | Categorical | No | 65.3; 34.7 |
| 6 | German | 1,000 | 21 | Mix | No | 70; 30 |
| 7 | Car-eval | 1,728 | 7 | Categorical | No | 22.3; 3.9; 70; 3.8 |
| 8 | Hypo | 3,163 | 26 | Mix | Yes | 95.2; 4.8 |
| 9 | kr-vs-kp | 3,196 | 37 | Categorical | No | 52.2; 47.8 |
| 10 | Waveform | 5,000 | 22 | Numeric | No | 33.2; 32.9; 33.9 |
| 11 | Mushroom | 8,124 | 23 | Categorical | Yes | 51.7; 48.3 |
| 12 | Nursery | 12,960 | 9 | Categorical | No | 33.3; 32.9; 31.2; 2.5; 0.01 |
| 13 | Adult | 48,842 | 14 | Mix | Yes | 76; 24 |
| 14 | Bank | 45,211 | 17 | Mix | No | 11.7; 88.3 |
| 15 | Skin | 245,057 | 4 | Numeric | No | 20.7; 79.3 |
| 16 | s-mushroom | 61,069 | 21 | Mix | Yes | 44.5; 55.5 |
| 17 | Connect-4 | 67,557 | 42 | Categorical | No | 65.8; 24.6; 9.6 |
| 18 | PUC-Rio | 165,632 | 19 | Mix | No | 28.6; 26.2; 7.5; 7.1; 30.6 |
| 19 | Census | 299,285 | 41 | Mix | Yes | 93.8; 6.2 |
| 20 | Gas-sensor-11 | 919,438 | 11 | Numeric | No | 32.9; 29.8; 37.3 |
| 21 | Gas-sensor-12 | 919,438 | 12 | Numeric | No | 32.9; 29.8; 37.3 |
| 22 | Cover-type | 581,012 | 55 | Mix | No | 36.4; 48.8; 6.2; 0.5; 1.6; 3; 3.5 |
| 23 | pamap2 | 1,942,872 | 33 | Numeric | Yes | 9.9; 6; 9.5; 5.4; 2.5; 9.8; 12.3; 9; 5.1; 12.3; 8.5; 9.7 |
| 24 | Susy | 5,000,000 | 19 | Numeric | No | 54.2; 45.8 |

We have performed a 10-fold cross-validation with a time-out of 72 hours for an algorithm execution on a dataset. For datasets *pamap2* and *susy*, the algorithms were only tested on the first fold of the 10-fold cross-validation. The same train-test splits are used for all algorithms. CMAR and PyIDS run on the same discretized datasets as Lord. The input order of the training examples for Lord* is kept as it is in each cross-validation fold. All experiments were run on two Xeon Quad-core CPUs X5570 @2.93GHz, hence eight cores, and 46 GB of available memory.

---
[7] The authors' source code can be found at https://github.com/jirifilip/pyIDS.
[8] https://www.philippe-fournier-viger.com/spmf/.
[9] https://cgi.csc.liv.ac.uk/~frans/KDD/Software/.

Besides the above rule learners, we also compared LORD with a black-box approach, i.e. SMO (Platt, 1998), Weka's sequential minimal optimization algorithm for training a SVM classifier. The default and the best settings on each data set were used for the comparison. We selected the best performance among multiple values of the complexity parameter $C$ for SMO. In terms of accuracy, LORD beats SMO on the larger datasets (12–23), but loses on most of the smaller datasets (7 out of 1–10), and sets a tie on mushroom dataset. With respect to runtime, there is not much difference between LORD and SMO on the small datasets since both run in lesser than 1 s, but the runtime on larger datasets grows enormously for SMO, e.g. SMO cannot complete its learning on susy dataset in the time-out setting. On average over the datasets (1–23), LORD is better than SMO for both runtime (in s) and accuracy, i.e. with the default setting, (354, 0.9297) for LORD vs. (7825, 0.8749) for SMO, and with the best settings, (354, 0.9339) vs. (21019, 0.8773).

## 4.2 Predictive accuracy

Table 2 shows the classification accuracy of the algorithms. LORD is implemented to execute in parallel for the rule discovery phase with the thread count equals the detected number of cores of a machine. It runs with a fixed/default m-estimate parameter setting of $m = 0.1$, as shown in column 3. We also show the results of the setting for which LORD achieved its best accuracy[10] and the setting for enhanced execution performance LORD* (Sect. 3.4) with the default $m = 0.1$ in column 5.

RIPPER is generally used with its default settings, but we tried different settings for the number of optimization runs $o$ with 0, and 2 (the default value), which are shown respectively in columns 6, and 7. This parameter is important, as these optimization runs have a large positive effect on RIPPER's accuracy, but are also quite expensive. CMAR in column 8 and CG in column 9 were used in their default settings recommended by the authors, and PYIDS with two settings using $k = 50$ and $k = 150$ association rules is respectively shown in columns 10 and 11. CG is for binary classification only, therefore in column 9 there are no results (//) for this algorithm on multi-class datasets.

In order to compare the algorithms, we group the values in Table 2 according to the basic algorithm, and highlight a value in **bold** if it outperforms the best values of other algorithms, but not necessarily other parameter settings of the same algorithm. This allows to also compare, e.g., the parameter setting of LORD with a fixed parameter $m = 0.1$ to all competitors, which would not be possible if we had only marked the best in each line, because LORD (best $m$) is always at least as good as LORD ($m = 0.1$). Because many experiments with the competitive algorithms could not be completed on the two largest datasets *pamap2* and *susy*, the average accuracy is derived from a main group of the first 22 datasets. Another average accuracy is calculated from a subgroup of datasets which can be processed successfully by CG. It's obvious that LORD gives the highest average rank and accuracy for all the three settings on the datasets winning 11/22 ($m = 0.1$), 13/22 (best $m$)

---

[10] We searched manually for the best $m$ value for each dataset, usually with increments of 1 in the range [0, 10] until a local optimum was found, which was then refined in a few 0.1 steps. On average, no more than 10 configurations were tested for each dataset in column 4.

**Table 2** The accuracy of algorithms performing on the datasets

| # | Datasets | LORD(m = 0.1) | LORD(best m) | LORD* (m = 0.1) | RIPPER(o = 0) | RIPPER(o = 2) | CMAR | CG | PyIDS (k =50) | PyIDS (k = 150) |
|---|---|---|---|---|---|---|---|---|---|---|
| 1 | Lymph | 0.8109 | **0.8309** | 0.8109 | 0.7095 | 0.7376 | 0.8166 | // | 0.5476 | 0.5476 |
| 2 | Wine | 0.9441 | **0.9555** | 0.9441 | 0.9382 | 0.9271 | 0.9500 | // | 0.8820 | 0.9437 |
| 3 | Vote | 0.9427 | 0.9473 | 0.9450 | 0.9518 | 0.9541 | 0.9426 | **0.9564** | 0.8826 | 0.9174 |
| 4 | Breast | 0.9556 | 0.9613 | 0.9570 | 0.9484 | 0.952 | **0.9642** | 0.9613 | 0.9527 | 0.9499 |
| 5 | Tic-tac-toe | 0.9874 | 0.9916 | 0.9874 | 0.9739 | 0.9718 | 0.9916 | **0.9958** | 0.7974 | 0.8131 |
| 6 | German | **0.7500** | **0.7510** | 0.7489 | 0.7210 | 0.7110 | 0.7390 | 0.7140 | 0.6970 | 0.6930 |
| 7 | Car-eval | 0.8998 | **0.9022** | 0.9004 | 0.8350 | 0.8599 | 0.7962 | // | 0.7696 | 0.8084 |
| 8 | Hypo | 0.9829 | 0.9845 | 0.9829 | **0.9911** | 0.9908 | 0.978 | 0.984 | 0.9522 | 0.9522 |
| 9 | kr-vs-kp | 0.9956 | **0.9959** | 0.9953 | 0.9896 | 0.9899 | 0.9364 | 0.9436 | 0.6877 | 0.7074 |
| 10 | Waveform | 0.7798 | 0.8119 | 0.7806 | 0.7736 | 0.7960 | **0.8216** | // | 0.3621 | 0.4472 |
| 11 | Mushroom | **1.0** | **1.0** | **1.0** | 0.9998 | **1.0** | 0.9943 | 0.9964 | 0.9304 | 0.9341 |
| 12 | Nursery | 0.9849 | 0.9852 | 0.9851 | 0.9614 | 0.9709 | 0.8959 | // | 0.7472 | 0.7445 |
| 13 | Adult | 0.8520 | **0.8559** | 0.8513 | 0.8365 | 0.8450 | 0.8335 | 0.8256 | 0.7713 | 0.7740 |
| 14 | Bank | 0.8949 | 0.8959 | 0.8928 | 0.8938 | **0.8994** | 0.8830 | 0.8928 | 0.8829 | 0.8831 |
| 15 | Skin | 0.9971 | 0.9971 | 0.9970 | **0.9991** | **0.9991** | 0.8892 | 0.9611 | 0.7924 | 0.7982 |
| 16 | Sec-mushroom | 0.9993 | **0.9993** | 0.9992 | 0.9969 | 0.9978 | 0.9239 | 0.9131 | 0.6041 | 0.7213 |
| 17 | Connect-4 | 0.8186 | **0.8206** | 0.8177 | 0.7280 | 0.7540 | 0.6701 | // | 0.6584 | 0.6584 |
| 18 | Puc-rio | 0.9534 | 0.9567 | 0.9533 | 0.9829 | **0.9869** | 0.7800 | // | 0.4445 | 0.4957 |
| 19 | Census | 0.9514 | 0.9514 | 0.9516 | 0.9481 | 0.9506 | 0.9379 | Error | 0.9379 | 0.9379 |
| 20 | Gas-sensor-11 | 0.9767 | **0.9771** | 0.9768 | 0.9255 | 0.9474 | 0.1884 | // | 0.3904 | 0.4100 |
| 21 | Gas-sensor-12 | 0.9965 | 0.9965 | 0.9965 | 0.9992 | **0.9994** | 0.1972 | // | 0.3912 | 0.4060 |
| 22 | Cover-type | 0.9152 | **0.9161** | 0.9118 | 0.8564 | 0.8943 | 0.5928 | // | 0.4882 | 0.4892 |
| 23 | Pamap2 | 0.9955 | 0.9963 | 0.9953 | OOM | OOM | 0.2657 | // | OOM | OOM |
| 24 | Susy | 0.7773 | 0.7860 | 0.7669 | OOM | OOM | 0.7477 | OOT | 0.5488 | 0.5566 |
| | Avg. acc. (3–6, 8–9, 11, 13–16) | 0.9416 | **0.9436** | 0.9415 | 0.9365 | 0.9374 | 0.916 | 0.9222 | 0.8137 | 0.8312 |
| | Avg. acc. (1–22) | 0.9268 | **0.9311** | 0.9266 | 0.9073 | 0.9152 | 0.8056 | // | 0.7077 | 0.7287 |
| | Avg. ranks (1–22) | 3.14 | **1.84** | 3.3 | 4.48 | 3.59 | 5.2 | // | 7.57 | 6.89 |

*OOM* Out of memory, *OOT* Out of time

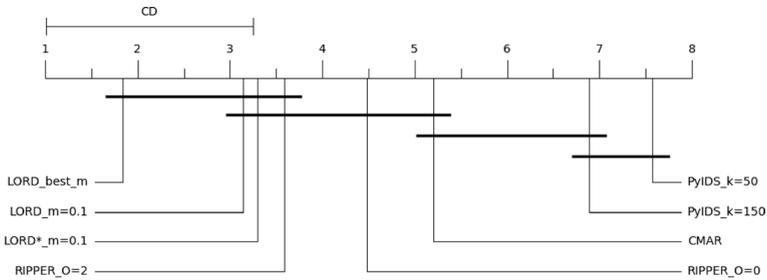

**Fig. 2** Nemenyi test on accuracy of algorithms on the first 22 datasets, $CD = 2.24$, $\alpha = 0.05$

and 11/22 (LORD*) over the competitors. For *pamap2* and *susy*, the performance of LORD is also superior to the competitors who could complete these tasks. The differences in average accuracy among the algorithms grows when moving from the subgroup to the main group of datasets. For example, the performance differences of the three settings of LORD compared to the second best RIPPER with ($o = 2$) increase from [0.42%, 0.62%, 0.41%] to [1.16%, 1.59%, 1.14%], and even more for the other algorithms.

The second best algorithm, RIPPER, wins 4/22 and 6/22 for its two settings. CG and CMAR both winning 2/22 respectively come at the third and the fourth place, and PYIDS with no wins positions at the last place. On average, PYIDS achieves a higher accuracy for larger values of $k$, but this is not consistent across all datasets (e.g., *breast*, *german*).

In order to assess whether these differences are statistically significant, we do a Friedman test (Demšar, 2006) based on the average ranks on the first 22 datasets. CG is not ranked because of its incompleteness for many datasets. The result shows a significant difference ($p$-value $= 2.027e - 18$) in accuracy, indicating that the null hypothesis that all algorithms have the same performance can be confidently rejected. The post-hoc Nemenyi test is visualized in Fig. 2 with the critical distance $CD = 2.24$ at significance level 0.95. It can be seen that the highest accuracy group includes LORD with the three settings and RIPPER ($o = 2$) in which the accuracy of LORD (best $m$) is consistently higher than RIPPER ($o = 2$) but not enough to indicate a significant difference by the test. LORD (best $m$) is significantly more accurate than the remaining algorithms, RIPPER without its optimization runs ($o = 0$), CMAR, and PYIDS.

### 4.3 Runtime comparison

Table 3 shows the average (derived from 10-fold cross-validation) runtime of all algorithms on each of the datasets. The last lines show the average of these values and the average rank of each method (after rounding the runtime at precision of 0.1 s) for the first 22 datasets. It is obvious that PYIDs is the slowest algorithm, and that its runtime increases fast with the number of considered rules $k$. On the small datasets, LORD and RIPPER run very fast, in less than 1 s, while CMAR and CG typically take a bit longer.

On the larger datasets, LORD is much faster than both RIPPER and CG. Without the optimization runs, RIPPER ($m = 0$) is faster than LORD in some cases but slower than LORD* ($m = 0.1$) for all the datasets. The optimization of RIPPER is necessary to improve its accuracy, and this makes RIPPER ($o = 2$) considerably slower than LORD in all the cases. Note that the average accuracy and rank of LORD* ($m = 0.1$) are also higher than that of RIPPER ($o = 2$). We can see that LORD* runs faster than LORD, especially much faster for the last

**Table 3** The runtime (in s) of algorithms performing on the datasets

| # | Datasets | LORD(m = 0.1) | LORD(best m) | LORD* (m = 0.1) | RIPPER(o = 0) | RIPPER(o = 2) | CMAR | CG | PYIDS (k = 50) | PYIDS (k = 150) |
|---|---|---|---|---|---|---|---|---|---|---|
| 1 | Lymph | 0.02 | 0.02 | 0.02 | 0.01 | 0.03 | 13.2 | // | 9.1 | 1093 |
| 2 | Wine | 0.03 | 0.03 | 0.02 | 0.01 | 0.02 | 52.1 | // | 11.6 | 2641 |
| 3 | Vote | 0.04 | 0.05 | 0.02 | 0.01 | 0.02 | 29.3 | 6.4 | 12.5 | 1482 |
| 4 | Breast | 0.04 | 0.04 | 0.03 | 0.02 | 0.05 | 0.2 | 9.3 | 126 | 3465 |
| 5 | Tic-tac-toe | 0.06 | 0.07 | 0.03 | 0.02 | 0.05 | 0.14 | 7.1 | 10.1 | 705 |
| 6 | German | 0.11 | 0.11 | 0.07 | 0.08 | 0.13 | 9.5 | 36.5 | 9.9 | 876 |
| 7 | Car-eval | 0.06 | 0.06 | 0.03 | 0.05 | 0.19 | 0.05 | // | 10.6 | 1195 |
| 8 | Hypo | 0.16 | 0.18 | 0.08 | 0.05 | 0.11 | 10.9 | 14.4 | 33.2 | 5841 |
| 9 | kr-vs-kp | 0.65 | 0.65 | 0.14 | 0.07 | 0.23 | 8.7 | 5.7 | 70.2 | 2408 |
| 10 | Waveform | 0.64 | 0.59 | 0.42 | 0.48 | 1.5 | 4.5 | // | 17.1 | 802 |
| 11 | Mushroom | 0.36 | 0.36 | 0.09 | 0.07 | 0.19 | 70.5 | 24.8 | 56.9 | 3896 |
| 12 | Nursery | 0.58 | 0.58 | 0.23 | 0.49 | 13.9 | 0.32 | // | 20.8 | 1776 |
| 13 | Adult | 4.8 | 4.8 | 1.5 | 2.6 | 16.0 | 32.5 | 285 | 75.5 | 853 |
| 14 | Bank | 10.5 | 10.5 | 1.7 | 1.3 | 7.7 | 146 | 606 | 74.1 | 1093 |
| 15 | Skin | 1.1 | 1.1 | 0.79 | 7.2 | 150 | 0.8 | 2941 | 340 | 1919 |
| 16 | Sec-mushroom | 4.0 | 4.0 | 0.87 | 6.8 | 174 | 14.9 | 5619 | 94.6 | 1142 |
| 17 | Connect-4 | 67.0 | 67.9 | 35.5 | 27.8 | 551 | 81.2 | // | 107 | 993 |
| 18 | Puc-rio | 33.7 | 33.5 | 22.1 | 111 | 1578 | 239 | // | 250 | 1335 |
| 19 | Census | 378 | 378 | 65.5 | 108 | 868 | 465 | Error[a] | 638 | 8223 |
| 20 | gas-sensor-l1 | 49.6 | 51.9 | 20.5 | 233 | 1617 | 3.7 | // | 1379 | 4891 |
| 21 | Gas-sensor-l2 | 56.2 | 57.3 | 21.2 | 498 | 7129 | 4.1 | // | 1341 | 5432 |
| 22 | cover-type | 1460 | 1480 | 523 | 6528 | 178034 | 1366 | // | 1357 | 4449 |
| 23 | pamap2 | 6063 | 6044 | 386 | OOM | OOM | 50.4 | // | OOM | OOM |
| 24 | susy | 52592 | 51218 | 15350 | OOM | OOM | 97.4 | OOT | 9435 | 29,109 |
| | Avg. runtime (1–22) | 94 | 95.1 | 31.5 | 342 | 8642.8 | 116 | // | 274.7 | 2568.6 |
| | Avg. ranks (1–22) | 3.5 | 3.75 | 1.73 | 2.95 | 5.09 | 4.89 | // | 6.27 | 7.82 |

*OOM* Out of memory, *OOT* Out of time

[a]The error happens while binarizing attributes by 'fit' function of class 'FeatureBinarizer' in the CG algorithm source code by the authors

three largest datasets, without losing much of the predictive accuracy of its counterpart LORD ($m = 0.1$).

For the average runtime over the first 22 datasets, LORD and LORD* are obviously the fastest ones, but over all 24 datasets, CMAR is faster than LORD* and LORD. This change comes mainly from the long runtime of LORD and LORD* on the *susy* dataset. CMAR runs the fastest for some large datasets (*gas-sensor-11*, *gas-sensor-12*, *pamap2*, and *susy*), but its accuracy on these datasets is much lower than that of LORD*. The reason is that these datasets are very sparse resulting in a low amount of generated frequent patterns and class association rules. Moreover, the chosen implementation LUCS-KDD of CMAR applies additional limitations on the number of found frequent patterns, rules and their length. The original version by Li et al. (2001) in which FP-GROWTH is used to discover frequent patterns, as implemented in SPMF, can result in long runtimes or memory overflow caused by large numbers of conditional pattern trees for dense datasets mined at the default minimum support 0.01, e.g. *census*, *connect-4*, *kr-vs-kp*.

In summary, LORD* is the fastest followed by RIPPER ($o = 0$), LORD, CMAR, RIPPER ($o = 2$) and PYIDS. Thus, LORD and LORD* find a better balance between accuracy and runtime than the other competitors.

### 4.4 Rule complexity

In terms of the rule complexity reported in Table 4, without considering CG, PYIDS is the best algorithm; however, for this advantage it sacrifices too much classification performance. RIPPER comes at the second place with a much better balance between the rule complexity and performance. The rule sets by LORD and LORD* are larger than those found by RIPPER, CG, and PYIDS but can be competitive to those by CMAR. For example, the rule sets found by LORD* are even smaller than those by CMAR for 14 out of the 24 datasets. Although the average rule lengths by LORD are shorter than those of RIPPER for some large datasets, the sizes of the found rule sets are, again, orders of magnitude larger than those of RIPPER. This was to be expected because LORD searches for a locally optimal rule for each training example, and thus, its rule sets are likely to contain many groups of rule variants and are typically much larger than sparse rule sets learned by conventional rule learners. However, the more complex rule sets by LORD are often compensated by more accurate classifications.

### 4.5 Influence of discretization

A possible trivial reason for the performance difference between LORD and the other algorithms on mixed or numerical datasets could be that LORD uses pre-discretized data (as do CMAR and PYIDS), whereas RIPPER and CG use their own internal discretization. In order to investigate this possibility, we also performed experiments where RIPPER and CG were used on the same discretized datasets as LORD.

Table 5 reports the results. Each tabular cell shows the average execution time (in s) on top and the average accuracy below. The values in parentheses show the corresponding values on the original datasets, which we duplicate here for convenience. The better values for the discretized datasets are highlighted in bold. After the discretization, the accuracy of RIPPER decreases for most of the datasets, except for *adult* and *bank* with ($o = 0$) and *sec-mushroom*. Similarly, CG loses accuracy for 5 out of 7 datasets, and gains for *adult* and *hypo*. For *census*, it results in a similar error message as on the

**Table 4** The rule complexity of algorithms performing on the datasets, rule count on the left and average rule length on the right of each cell

| # | Datasets | Lord (m = 0.1) | | Lord (best m) | | Lord* (m = 0.1) | | Ripper (o = 0) | | Ripper (o = 2) | | CMAR | | CG | | PyIDS (k = 50) | | PyIDS (k = 150) | |
|---|---|---|---|---|---|---|---|---|---|---|---|---|---|---|---|---|---|---|---|
| 1 | Lymph | 33.4 | 2.58 | 33.4 | 2.58 | 32.9 | 2.56 | 6.3 | 1.48 | 6.1 | 1.46 | 189.6 | 3.66 | // | | 16.7 | 1.95 | 51.5 | 1.97 |
| 2 | Wine | 16.9 | 1.93 | 16.7 | 1.97 | 16.9 | 1.93 | 4.1 | 1.28 | 3.9 | 1.26 | 118.0 | 2.55 | // | | 16.0 | 1.93 | 49.1 | 1.97 |
| 3 | Vote | 39.1 | 3.08 | 39.1 | 3.05 | 37.4 | 3.08 | 3.4 | 1.45 | 2.3 | 0.73 | 245.3 | 3.79 | 3.2 | 2.42 | 16.0 | 2.00 | 53.1 | 2.00 |
| 4 | Breast | 56.2 | 2.60 | 54.6 | 2.58 | 52.2 | 2.57 | 6.1 | 1.78 | 5.5 | 1.67 | 328.2 | 3.31 | 3.4 | 2.81 | 17.5 | 2.87 | 50.8 | 2.87 |
| 5 | Tic-tac-toe | 29.1 | 3.78 | 28.7 | 3.27 | 28.0 | 3.73 | 11.2 | 2.99 | 10.3 | 2.88 | 250.1 | 4.03 | 8.2 | 3.20 | 17.4 | 3.00 | 51.7 | 3.00 |
| 6 | German | 296.2 | 3.38 | 296.2 | 3.39 | 289.5 | 3.37 | 6.5 | 2.73 | 4.2 | 1.86 | 1784.7 | 3.80 | 5.4 | 3.70 | 15.9 | 1.99 | 49.7 | 2.00 |
| 7 | Car-eval | 220.3 | 4.82 | 220.3 | 4.82 | 208.3 | 4.80 | 31.6 | 3.93 | 32.5 | 3.72 | 178.8 | 2.40 | // | | 17.6 | 2.18 | 49.5 | 2.73 |
| 8 | Hypo | 119.4 | 2.41 | 127.3 | 2.38 | 108.9 | 2.36 | 3.5 | 1.72 | 2.4 | 1.29 | 300.4 | 3.47 | 3.9 | 1.25 | 17.5 | 1.84 | 51.0 | 1.96 |
| 9 | kr-vs-kp | 48.3 | 4.62 | 48.9 | 4.53 | 45.7 | 4.68 | 15.9 | 3.21 | 15.8 | 2.99 | 232.9 | 2.95 | 2.4 | 2.65 | 18.9 | 1.98 | 49.3 | 1.98 |
| 10 | Waveform | 1428.2 | 3.21 | 1065.4 | 2.93 | 1391.3 | 3.21 | 26.2 | 5.02 | 27.8 | 4.49 | 2014.5 | 2.79 | // | | 16.9 | 2.00 | 53.0 | 2.00 |
| 11 | Mushroom | 26.4 | 1.71 | 26.4 | 1.71 | 26.2 | 1.71 | 8.9 | 1.39 | 8.5 | 1.43 | 202.6 | 3.04 | 3.0 | 1.66 | 17.3 | 1.91 | 51.4 | 1.97 |
| 12 | Nursery | 572.2 | 5.28 | 571.7 | 5.29 | 565.1 | 5.26 | 122.2 | 4.29 | 121.7 | 4.56 | 442.4 | 3.19 | // | | 14.4 | 1.98 | 50.1 | 1.95 |
| 13 | Adult | 8213.0 | 4.66 | 7614.3 | 4.68 | 3360.0 | 4.57 | 28.1 | 5.79 | 16.7 | 4.28 | 6853.7 | 4.58 | 2.0 | 2.55 | 15.6 | 1.97 | 50.6 | 1.98 |
| 14 | Bank | 6666.0 | 4.67 | 6643.2 | 4.69 | 3245.1 | 4.52 | 14.6 | 3.99 | 13.8 | 3.07 | 12413.2 | 4.26 | 1.0 | 1.00 | 16.7 | 2.00 | 50.2 | 1.99 |
| 15 | Skin | 2258.4 | 2.31 | 2238.5 | 2.30 | 2047.1 | 2.33 | 39.9 | 5.36 | 36.5 | 5.33 | 133.4 | 1.58 | 2.0 | 4.00 | 15.9 | 2.00 | 51.8 | 2.00 |
| 16 | Sec-mushroom | 478.6 | 2.82 | 478.6 | 2.82 | 473.6 | 2.78 | 95.7 | 3.76 | 85.2 | 3.64 | 2144.1 | 3.65 | 19.1 | 2.71 | 15.9 | 1.90 | 52.0 | 1.95 |
| 17 | Connect-4 | 14341.1 | 6.41 | 14396.6 | 6.41 | 13370.9 | 6.31 | 138.4 | 6.39 | 128.8 | 5.45 | 2376.0 | 2.98 | // | | 15.2 | 2.00 | 49.6 | 2.00 |
| 18 | Puc-rio | 11812.9 | 3.14 | 10223.8 | 3.18 | 11416.4 | 3.12 | 238.8 | 6.13 | 243.8 | 6.02 | 6035.4 | 2.93 | // | | 16.8 | 2.00 | 51.6 | 2.00 |
| 19 | Census | 19230.4 | 5.23 | 19230.4 | 5.23 | 4455.4 | 4.61 | 53.9 | 6.96 | 48.3 | 5.97 | 5011.6 | 2.92 | Error | | 19.6 | 1.95 | 49.5 | 1.98 |
| 20 | Gas-sensor-11 | 20224.5 | 2.07 | 19878.0 | 2.08 | 18042.4 | 2.07 | 72.3 | 6.80 | 73.1 | 6.42 | 22.0 | 1.23 | // | | 17.2 | 1.90 | 51.1 | 1.96 |
| 21 | Gas-sensor-12 | 17174.2 | 2.12 | 17071.7 | 2.13 | 16324.6 | 2.12 | 180.5 | 5.92 | 181.8 | 5.89 | 23.8 | 1.25 | // | | 15.6 | 1.88 | 45.5 | 1.97 |
| 22 | Cover-type | 81308.8 | 5.02 | 81447.3 | 5.01 | 64408.9 | 5.00 | 1192.2 | 8.28 | 1385 | 7.89 | 1491.6 | 2.92 | // | | 19.1 | 1.99 | 47.4 | 1.98 |
| 23 | Pamap2 | 16827 | 3.07 | 14137 | 3.09 | 15824 | 3.05 | OOM | | OOM | | 486 | 2.54 | // | | OOM | | OOM | |
| 24 | Susy | 1611856 | 4.30 | 1,201,338 | 4.10 | 976,522 | 4.30 | OOM | | OOM | | 637 | 1.40 | OOT | | 18 | 2.0 | 63 | 2.0 |
| | Avg. values (1–22) | 8390.6 | 3.54 | 8261.4 | 3.5 | 6361.2 | 3.49 | 104.6 | 4.12 | 111.5 | 3.74 | 1945.1 | 3.06 | // | | 16.8 | 2.06 | 50.4 | 2.1 |
| | Avg. ranks (1–22) | 6.82 | 5.86 | 6.39 | 5.82 | 5.25 | 4.95 | 2.73 | 5.16 | 2.23 | 3.95 | 6.45 | 4.86 | // | | 1.91 | 2.45 | 4.23 | 2.93 |

*OOM* Out of memory, *OOT* Out of time

Machine Learning

**Table 5** Experimental results of RIPPER and CG algorithms on mixed and numerical datasets discretized by FUSINTER. The numbers in parentheses repeat the evaluations on the original datasets

| # | Datasets | RIPPER (o = 0) | | RIPPER (o = 2) | | CG | |
|---|---|---|---|---|---|---|---|
| 2 | Wine | (0.01) (0.9382) | 0.12 0.8611 | (0.02) (0.9271) | 0.21 0.9111 | // | // |
| 4 | Breast | (0.02) (0.9484) | 0.14 0.9457 | (0.05) (0.9527) | 0.14 0.9428 | (9.3) (0.9613) | **6.9** 0.9527 |
| 6 | German | (0.08) (0.721) | 0.17 0.708 | (0.13) (0.711) | 0.28 0.697 | (36.5) (0.714) | **36.0** 0.706 |
| 8 | Hypo | (0.05) (0.9911) | 0.19 0.9873 | (0.11) (0.9908) | 0.29 0.9873 | (14.4) (0.9841) | 22.0 **0.9898** |
| 10 | Waveform | (0.48) (0.7736) | **0.42** 0.7190 | (1.49) (0.7962) | **1.2** 0.7412 | // | // |
| 13 | Adult | (2.6) (0.8365) | 3.7 **0.8404** | (16.0) (0.8451) | 28.4 0.8424 | (285) (0.8256) | **278** **0.8496** |
| 14 | Bank | (1.3) (0.8938) | 1.9 **0.8973** | (7.7) (0.8994) | 11.5 0.8982 | (606) (0.8928) | **433** 0.8928 |
| 15 | Skin | (7.2) (0.9991) | 40.3 0.9895 | (150) (0.9991) | 10360 0.9910 | (2941) (0.9611) | 4033 0.9296 |
| 16 | Sec-mushroom | (6.8) (0.9969) | 8.7 **0.9987** | (174) (0.9978) | 326 **0.9991** | (5619) (0.9131) | 7019 0.8692 |
| 18 | Puc-rio | (111) (0.9829) | 229 0.9017 | (1578) (0.9869) | 12245 0.9427 | // | // |
| 19 | Census | (108) (0.9481) | 125 0.9465 | (868) (0.9506) | **655** 0.9489 | Error | Error |
| 20 | Gas-sensor-11 | (233) (0.9255) | 1223 0.7222 | (1617) (0.9474) | OOM | // | // |
| 21 | Gas-sensor-12 | (498) (0.9992) | OOM | (7129) (0.9994) | OOM | // | // |
| 22 | Cover-type | (6528) (0.8564) | 12,248 0.6716 | (178034) (0.8943) | OOM | // | // |

*OOM* Out of memory

original dataset. CG runs slightly faster for some discretized datasets but much slower for *skin* and *sec-mushroom*. RIPPER also takes a longer runtime for most of the discretized datasets. In 4 cases, it also ran out of memory (in particular in the default setting with 2 optimization runs).

Table 6 shows the runtime of the LORD algorithm including the data discretization time compared to the runtime of RIPPER and CG on 4 datasets. For the other datasets, the discretization time is negligible, typically some tens to hundred milliseconds. In comparison to RIPPER's default configuration with two optimization runs, LORD still does not lose its advantage in execution time with the additional discretization time.

In summary, discretization does not generally improve the performance of CG and RIPPER, so that we can conclude that the discretization does not provide an unfair advantage to LORD. Note that we also have not spent much effort on optimizing the discretization.

**Table 6** Data discretization time (in s) included in the runtime of LORD and LORD* compared to others'

| # | Datasets | LORD (m = 0.1) | LORD (best m) | LORD* (m = 0.1) | RIPPER (o = 0) | RIPPER (o = 2) | CG |
|---|---|---|---|---|---|---|---|
| 20 | *gas-sensor-11* | 628 | 630 | 598 | 233 | 1617 | // |
| 21 | *gas-sensor-12* | 1929 | 1930 | 1894 | 498 | 7129 | // |
| 23 | *pamap2* | 53291 | 53272 | 47614 | OOM | OOM | // |
| 24 | *Susy* | 79488 | 78114 | 42246 | OOM | OOM | OOT |

*OOM* Out of memory, *OOT* Out of time

### 4.6 Scalability analysis

This section analyzes the scalability of LORD which is potentially affected by two factors, data size and the number of threads running in the rule learning phase. For these experiments, we use the *susy* dataset with subsets increasing in size, up to 5,000,000 examples. Figure 3a and b respectively show the memory consumption and the runtime of LORD w.r.t. the number of examples. While the memory consumed in the first phase (before the rule learning phase) and by the data structures increases approximately linearly with the data size, the total memory peak and the runtime show a super-linear increase. The runtime differences between 1 and 2 million data points is approximately 2.17 times smaller than the corresponding increase between 4 and 5 million. For data sizes from 1 to 3 million, the total memory peak is the memory peak in the rule learning phase which is greater than that in the first phase; but for data sizes from 4 to 5 million, the total memory peak is the memory peak in the first phase, where in addition to the PPC-tree, we also need to store N-lists of selectors and re-code the training examples as arrays of selector IDs.

In the second phase, the memory for PPC-trees is freed, and the memory required for storing the N-lists of selectors and the R-tree is much smaller and also increases at a slower rate than the memory consumed by the PPC-tree. The main reason for this is that the N-nodes in N-lists contain only a PP-code and the corresponding frequency (Definition 5), whereas the PPC-nodes additionally store a *selector*, a children list and a reference to a parent to maintain the tree structure (Definition 4). This also allows for a more efficient implementation as a 2-dimensional array ($3 \times l$), where each of the three components of the $l$ N-nodes in the list is stored in a separate dimension.

The memory used by the data structures (i.e., PPC-tree, N-lists of selectors, R-tree) only depends on the data size but not on the number of threads for learning rules. Even though there is only a single master copy of distinct selectors, which is read-only and shared among the threads, each rule learning thread maintains a local *NListSet* for caching the N-lists of selector-sets that it encounters while finding the best rule for each single example. Thus, the memory peak used in the rule learning phase increases w.r.t. the number of rule-finding threads and may eventually overtake the consumed memory peak used in the first phase. We verify this assumption with experiments on the full *susy* dataset shown in Fig. 3c. The memory peak used in the learning phase increases linearly with the number of threads, but remains smaller than the memory occupied by the PPC-tree. Therefore, for a modest number of threads (we could experiment with up to 8 threads), the total consumed memory peak used by LORD is the memory peak during the construction of the PPC-tree and N-lists of selectors, which does not depend on the number of threads.

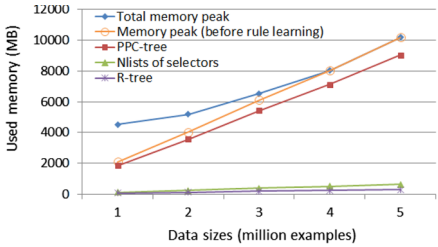
(a) Consumed memory w.r.t. example counts of the *susy* dataset, default setting ($m$=0.1, 8 threads)

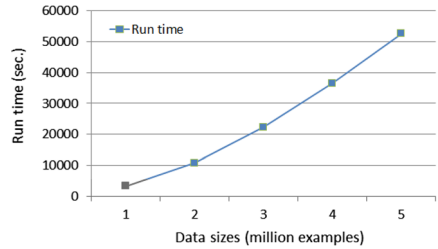
(b) Runtime w.r.t. example counts of the *susy* dataset, default setting ($m$=0.1, 8 threads)

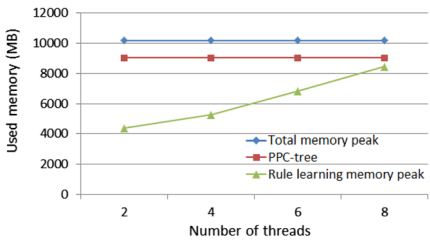
(c) Consumed memory w.r.t. thread counts, running on the full *susy* dataset, $m$=0.1

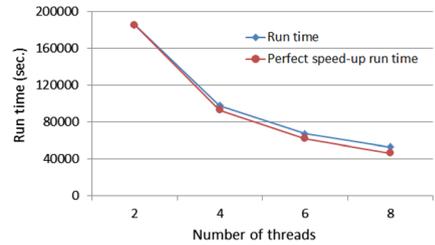
(d) Runtime w.r.t. thread counts, running on the full *susy* dataset, $m$=0.1

**Fig. 3** Influence of data sizes and thread counts on the consumed memory and the runtime

Figure 3d shows the runtime of LORD on the full *susy* dataset w.r.t. the number of threads. We can see that it is very close to an optimal speed-up, which would be achieved if increasing the number of threads by a factor of 2 roughly reduces the runtime by a factor of 2. This comes from the fact that the single-thread construction of the PPC-tree and N-lists runs comparably fast, so that the parallel rule learning phase takes most of the runtime of the LORD algorithm. This, in turn can run in parallel with very high load balance thanks to the inherent independence of finding the best rule for each single example.

In summary, it can be seen that LORD can confront memory-related scalability as efficiently as PPC-trees or similar tree structures such as FP-trees.

### 4.7 Influence of m-estimate Heuristic

Figure 4 shows the impact of parameter $m$ of the m-estimate heuristic on the classification accuracy performance on a representative subset of the 24 datasets from our main experiments. The optimal value range of the $m$ parameter can vary among datasets but all accuracy curves share the characteristic of a single peak value with a gradual reduction of the accuracy on both sides. This is not surprising, as it is known that the $m$-parameter provides a flexible trade-off between precision, which is known to overfit, and weighted relative accuracy, which is known to overgeneralize in predictive rule learning (Fürnkranz and Flach 2005; Janssen and Fürnkranz 2010). A special case is *mushroom* which is noise-free and thus has a flat peak, retaining a perfect accuracy of 1 for the value range [0, 10] of *m*. In general, we can see that the accuracy changes in simple shapes w.r.t the values of parameter *m*, hence it is easy to find the best value of *m*.

### 4.8 Analysis of hill-climbing variants

Recall from Sect. 3.4 that LORD's rule refinement process consists of a single greedy rule growing phase, followed by a single greedy rule pruning phase, as has also been realized in many other rule learning algorithms such as RIPPER. LORD thus finds a local optimum with respect to this 2-phase optimization procedure. However, one may rightfully argue,[11] that this local optimum could be further improved with additional pruning and growing steps, and that, in fact, a 2-phase local optimum may not necessarily be optimal in that case.

To check this, we have implemented OVERLORD, a variant that adds another greedy refinement phase in the opposite direction if the previous one has resulted in an improvement of the current local optimum. Thus, if the pruning phase of LORD improves the local optimum of the previous growing phase, OVERLORD follows up with another growing phase, which, in case it further improves the local optimum, is followed with another pruning phase, etc. The process terminates when the last phase did not achieve any improvement. As the phase before the last one has also not yielded further improvements into the opposite direction (which is why there was a switch in direction), the found rule is locally optimal w.r.t. both, specialization and generalization.

We have experimentally compared LORD and OVERLORD on all 24 datasets of Table 1. On 13 datasets, the results were exactly the same (except for minor differences in runtime). Table 7 summarizes the results in terms of accuracy, rule complexity, and runtime on the remaining 11 (medium to very large) datasets. We show the results of two settings, the default $m = 0.1$ and the best $m$ for each dataset. The better accuracy is highlighted in bold. The runtime of the two versions is similar for the first 9 datasets, but for the last 2 very big ones the additional runtime for OVERLORD is noticeable.

In cases where a difference is noticeable, it is typically less than 1%. Also, maybe somewhat surprisingly, it is not the case that the improved local optimum on the training data consistently improves the performance on the test data. For $m = 0.1$, we observed 4 wins for LORD and 5 wins for OVERLORD (with a total of 15 ties, including the 13 datasets that are not shown), and for the best $m$ on each dataset LORD was ahead on 6 datasets and OVERLORD on 4 datasets, with 14 ties. We explain this with a somewhat increased tendency towards overfitting, because the rule lengths of OVERLORD are typically slightly longer than those of LORD, in that the rules that are further refined after the first pruning phase, obviously become more complex, thus apply to fewer data points, and are rarely ever further improved with a second pruning phase. As this is a result of the search procedure, it may also be viewed as an instance of over-searching (Quinlan and Cameron-Jones, 1995; Janssen and Fürnkranz, 2009). In any case, the differences are only very small and occur without a clear pattern.

In general, we conclude from these experiments that two optimization phases, one for growing and one for pruning, are sufficient, and that additional phases can be applied as an optional setting for slightly better performance in some cases, but may also reduce the performance in others.

---

[11] We are indebted to one of our reviewers for this observation.

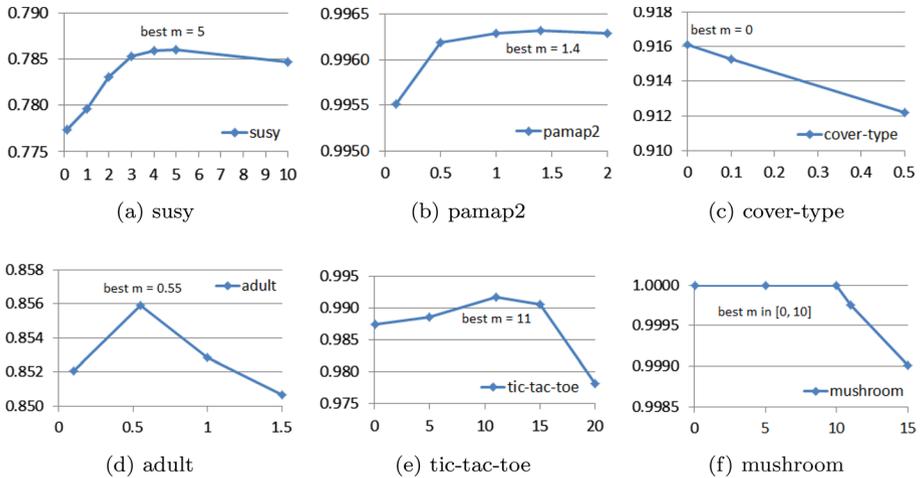

**Fig. 4** Influence of m-estimate heuristic on the classification accuracy. The x-axes and y-axes indicate the values of $m$ parameter of m-estimate heuristic and classification accuracy respectively

### 4.9 Analysis of rule filtering

Finally, we take a closer look at the high number of rules that are learned by LORD. In principle, LORD tries to identify the best rule for every single example. However, many of these rules are duplicates, already when they are found, and others are removed in a post-processing phase. We look at the magnitude of these reductions and evaluate whether such a high number of rules is necessary.

Columns 3 and 4 of Table 8 show the number of instances, and the number of best rules discovered from these instances. We see that on average, the number of rules is about 18% of the number of examples, which means that approximately five examples share the same best rule after our first greedy search phase. However, the exact values vary considerably, ranging from less than 1% in the easy *mushroom* dataset to more than half in the large *susy* dataset. This seems to relate to semantic or distributional properties of the dataset, as no clear relation to characteristics such as the size of the dataset is recognizable.

Furthermore, as mentioned in Sect. 3.4, LORD has an additional filtering phase which somewhat compacts the resulting rule sets by removing rules for all examples for which a better rule can be found in the learned set of rules. The last two columns of Table 8 show the effect of this strategy. The highest reduction happens for *pamap2*, where 86.77% of the learned rules are filtered out. On average, about 38% of the rules found in the learning phase will be eliminated in the filter phase, even though every rule has been a local optimum for at least one training example in the rule search phase. This means that there exists a large portion of training examples which abandon their own local best rule and adopt a local best rule of another training example as a better one. Thus, the negative effects of a greedy search for local optima are greatly reduced by this additional simple filter.

However, the remaining numbers of rules are still extremely large. For example, for the 5 million examples of *susy*, more than 2.8 million different rules are found in the first place, which are reduced to ca. 1.6 million rules after filtering. An obvious question is whether this large number of rules can be further reduced. While we cannot give a definite answer, we try to shed some light on this question by looking at the prediction quality with a varying rule quality threshold.

**Table 7** Comparison between LORD and OVERLORD on the 11 datasets, where performance differences were noticeable. Each cell shows (from top to bottom) the runtime (in s), the average accuracy, the number of generated rules, and the average rule length

| # | Datasets | m = 0.1 | | Best m | |
|---|---|---|---|---|---|
| | | LORD | OVERLORD | LORD | OVERLORD |
| 10 | waveform | 0.64 | 0.64 | 0.59 | 0.64 |
| | | 0.779800 | 0.779800 | **0.811999** | 0.810600 |
| | | 1428.2 | 1428.2 | 1065.4 | 1074.1 |
| | | 3.2111 | 3.2111 | 2.9377 | 2.9474 |
| 13 | adult | 4.8 | 5.1 | 4.8 | 5.2 |
| | | 0.852053 | **0.852135** | 0.855923 | **0.855943** |
| | | 8213.0 | 8216.1 | 7614.3 | 7618.6 |
| | | 4.6638 | 4.6657 | 4.6864 | 4.6898 |
| 14 | bank | 10.5 | 10.6 | 10.5 | 10.6 |
| | | **0.894937** | 0.894915 | 0.895932 | 0.895932 |
| | | 6666.0 | 6666.3 | 6643.2 | 6642.7 |
| | | 4.6726 | 4.6745 | 4.6988 | 4.7014 |
| 17 | connect-4 | 67.0 | 68.3 | 67.9 | 68.6 |
| | | 0.818627 | 0.818627 | **0.820625** | 0.820462 |
| | | 14341.1 | 14335.8 | 14396.6 | 14391.6 |
| | | 6.4189 | 6.4230 | 6.3927 | 6.3966 |
| 18 | puc-rio | 37.7 | 37.7 | 37.5 | 37.9 |
| | | **0.953427** | 0.953409 | **0.956729** | 0.956536 |
| | | 11812.9 | 11810.9 | 10223.8 | 10280.0 |
| | | 3.1432 | 3.1435 | 3.1842 | 3.1939 |
| 19 | census | 378 | 379 | 378 | 379 |
| | | 0.951447 | **0.951450** | 0.951447 | **0.951450** |
| | | 19230.4 | 19231.4 | 19230.4 | 19231.4 |
| | | 5.2372 | 5.2426 | 5.2372 | 5.2426 |
| 20 | gas-sensor-11 | 49.6 | 50.1 | 51.9 | 52.3 |
| | | **0.976782** | 0.976772 | **0.977138** | 0.977124 |
| | | 20224.5 | 20224.6 | 19878.0 | 19874.6 |
| | | 2.0794 | 2.0799 | 2.0899 | 2.0909 |
| 21 | gas-sensor-12 | 56.2 | 56.5 | 57.3 | 57.7 |
| | | **0.996519** | 0.996514 | **0.996548** | 0.996544 |
| | | 17174.2 | 17164.9 | 17071.7 | 17058.0 |
| | | 2.1239 | 2.1240 | 2.1305 | 2.1308 |
| 22 | cover-type | 1460 | 1493 | 1480 | 1497 |
| | | 0.915249 | **0.915375** | 0.916129 | **0.916258** |
| | | 81308.8 | 81338.2 | 81447.3 | 81491.6 |
| | | 5.027 | 5.0349 | 5.0191 | 5.0264 |
| 23 | pamap2 | 6063 | 7158 | 6044 | 7051 |
| | | 0.995517 | **0.995527** | 0.996314 | **0.996397** |
| | | 16827 | 16777 | 14137 | 14063 |
| | | 3.0705 | 3.0706 | 3.0926 | 3.1072 |
| 24 | susy | 52592 | 54002 | 51218 | 52447 |
| | | 0.777344 | **0.777346** | **0.786064** | 0.786026 |
| | | 1611856 | 1611940 | 1201338 | 1215561 |
| | | 4.3078 | 4.3079 | 4.1061 | 4.1147 |
| Avg. accuracy | | 0.901063 | **0.901079** | **0.905895** | 0.905752 |

Table 8 Reduction on the number of rules in the learning and the filtering phase

| # | Datasets | # Exs. | Original # Rules | Reduction (% of Exs.) | Kept # Rules | Reduction (% of Orig.) |
|---|---|---|---|---|---|---|
| 1 | lymph | 148 | 48 | 67.57 | 35 | 27.08 |
| 2 | wine | 178 | 18 | 89.89 | 16 | 11.11 |
| 3 | vote | 435 | 46 | 89.43 | 38 | 17.39 |
| 4 | breast-cancel | 699 | 70 | 89.99 | 58 | 17.14 |
| 5 | tic-tac-toe | 958 | 201 | 79.02 | 31 | 84.58 |
| 6 | german | 1000 | 472 | 52.80 | 298 | 36.86 |
| 7 | car-eval | 1728 | 233 | 86.52 | 218 | 6.44 |
| 8 | hypo | 3163 | 214 | 93.23 | 121 | 43.46 |
| 9 | kr-vs-kp | 3196 | 139 | 95.65 | 46 | 66.91 |
| 10 | waveform | 5000 | 2332 | 53.36 | 1363 | 41.55 |
| 11 | mushroom | 8124 | 33 | 99.59 | 26 | 21.21 |
| 12 | nursery | 12,960 | 752 | 94.20 | 591 | 21.41 |
| 13 | bank | 45,211 | 10,644 | 76.46 | 6584 | 38.14 |
| 14 | skin | 245,057 | 2310 | 99.06 | 2179 | 5.67 |
| 15 | sec-mushroom | 61,069 | 1377 | 97.75 | 476 | 65.43 |
| 16 | adult | 48,842 | 12,101 | 75.22 | 8507 | 29.70 |
| 17 | connect-4 | 67,557 | 24,921 | 63.11 | 14,404 | 42.20 |
| 18 | PUC-Rio | 165,632 | 20,217 | 87.79 | 11,884 | 41.22 |
| 19 | census | 299,285 | 35,976 | 87.98 | 19,241 | 46.52 |
| 20 | gas-sensor-11 | 919,438 | 30,669 | 96.66 | 20,422 | 33.41 |
| 21 | gas-sensor-12 | 919,438 | 28,712 | 96.88 | 17,290 | 39.78 |
| 22 | cover-type | 581,012 | 143,271 | 74.34 | 82,548 | 42.38 |
| 23 | pamap2 | 1,942,872 | 127,149 | 93.46 | 16,827 | 86.77 |
| 24 | susy | 5,000,000 | 2,880,375 | 42.40 | 1,611,856 | 44.04 |
| Average reduction | | | | 82.64 | | 37.93 |

Figure 5 depicts the classification performance of LORD against the percentage of best rules in rule sets for the 12 medium and big datasets.[12] Every data point in the figure shows the accuracy of a rule set consisting of the $p\%$ of the rules with the highest evaluation. Even though the increments in the first steps are larger than the increments in the later steps, we can see that in most cases, the accuracy of LORD continues to increase with increasing sizes of the rule sets, as can, e.g., be clearly observed for the two *gas-sensor* datasets, as well as for *cover-type* or *PUC-Rio*, the most notable exception being *census* and *bank*. This implies that even though better rules seem to contribute more to the overall predictive accuracy, the rules that contribute to improving the classification performance are distributed throughout the range of the ordered rule set, and that seemingly weaker rules are also necessary for maintaining a high classification performance.

---

[12] The same experiments on the small datasets give similar results with the same trend but the accuracy increase is not as smooth as that on the medium and big datasets because of fluctuations caused by the small numbers of test examples.

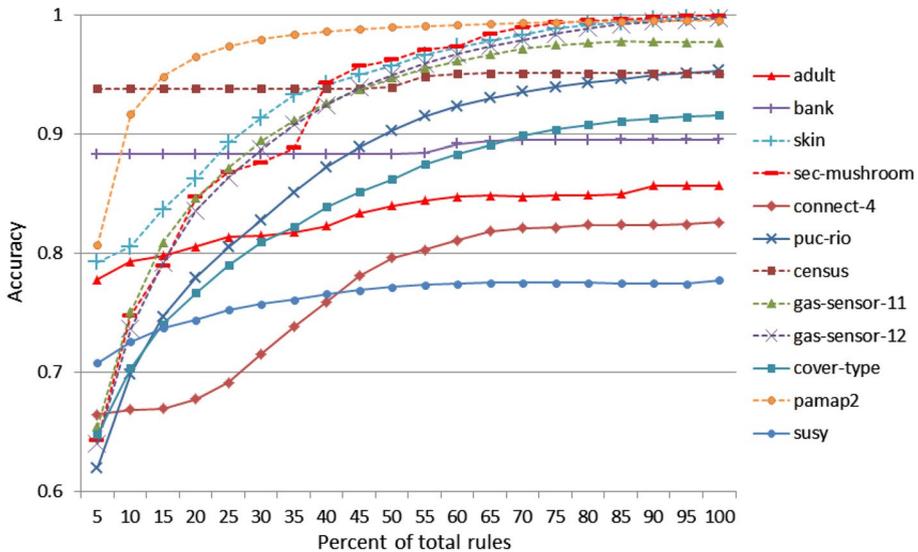

**Fig. 5** Classification performance of LORD against the percentage of best rules remained in rule sets produced from medium and big datasets

This result seems to reflect that higher quality rules which recognize usual and common cases are not enough to discriminate all cases well. Lower quality rules (with low coverage and in large quantity) are responsible for recognizing rare and exceptional cases that should also be in rule sets for better classification accuracy. This is reminiscent of the result by Holte et al. (1989), who observed that small disjuncts (i.e., rules with a low coverage) make up a fair percentage of the overall accuracy of a classifier, and that their removal may be futile.

In any case, we note that the number of covering rules for an unseen example is much smaller than the total number of rules in the rule sets, and the examination of the set of rules that cover an example is much easier and more focused than for the complete set of rules. When it comes to interpretation, one only has to deal with a small percentage of best rules which recognize popular cases for general principles. In this way, the complexity of interpreting rule sets can be reduced. In fact, unseen examples are always classified with a single rule (cf. Definition 8), which yields a natural justification for the predicted class. Nevertheless, clearly, interpretability is not the main focus of the LORD rule learner.

## 5 Conclusion

With the LORD algorithm, we have introduced a new approach to predictive rule learning in which a locally optimal rule is found for each training example, and the best covering rule of a new example is chosen for the classification. Although a large percentage of training examples share their local optima after a final filtering phase, the number of rules discovered by LORD is larger than that by conventional rule learners. Nevertheless, the rule sets learned by LORD outperform state-of-the-art rule learners in terms of predictive accuracy. More importantly, despite the large number of found rules, it is very efficient and can be applied to very large databases that cause problems for its competitors. This is primarily due to the adoption of efficient data structures, which have been developed in association rule discovery, to a classification learning setting. LORD also features an inherently parallel search for rules, which could be transformed to a massively parallel architecture for dealing with big data.

We have also seen that even though better rules contribute more to the predictive accuracy, the rules contributing on the improvement of classification performance are distributed throughout the range of an ordered rule set, so that a further filtering of the rules appears to be non-trivial. Although the remaining rules in a rule set after the filter step are a local optimum for at least one training example, it is still possible that rules can be removed without changing the performance on a given test set. A closer investigation of this issue, as well as a method for detecting such redundant rules will help to reduce the rule set complexity, and improve the interpretability of the found rule sets.

So far, we have not given much thought to the classification phase of the algorithm. Currently, we use the best (according to the training set performance) rule among all rules that cover a test example. Other rule selection methods, or possibly also voting techniques might further improve the prediction performance. In particular, we are also considering to adapt the method to a transductive setting, where the best rule for each *test* example is learned on the fly and—if it is a new local optimum—added to the rule set. However, this problem turns out to be harder than expected, and a straight-forward adaptation of LORD, which for each possible class forms the best rule covering the test example, performed worse than LORD.

## Appendix: Lemma for Property 3

**Lemma 1** *The N-list of a k-selector-set generated with Definition 7 (ii) is identical to the N-list of the k-selector-set generated with Definition 7 (i)*

**Proof** We prove the lemma by induction over the selector-set length *k*:

1. For $k = 2$, the two N-lists of 2-selector-set $S = s_1 s_2$ calculated based on Definition 7 (i) and (ii) are trivially identical because they are both generated from the same two N-lists of single selectors $s_1$ and $s_2$.
2. Assume that the lemma holds for $k - 1$, i.e., both, Definition 7 (i) and (ii), generate the same N-list $NL_{k-1}$ for the $(k-1)$-selector-set $S = s_1 s_2 \cdots s_{k-1}$. So we have:

    (*) that $NL_{k-1}$ contains N-nodes (associated with selector $s_{k-1}$) which convey information about the number of all paths (from the tree root to leaves) which contain all the selectors in the sequence $s_1, s_2, ..., s_{k-1}$ going up to the root.
3. We need to prove that the lemma also holds for *k*. In other words, we prove that $NL_k$ calculated from $NL_{k-1}$ and N-list $NL$ of $(k-1)$-selector-set $s_1 s_2 \cdots s_{k-2} s_k$ via Equation (2) and $NL'_k$ calculated from $NL_{k-1}$ and N-list $NL'$ of selector $s_k$ via Equation (2) are identical. We have:

(**) that $NL_k$ is $NL'_k$ or a subset of $NL'_k$ because they both share the same input N-list $NL_{k-1}$ in the calculation method (2), and $NL$ is a subset of $NL'$.

(***) that $NL'_k$ contains N-nodes (associated with selector $s_k$) which convey information about the number of all paths which contain all the selectors in sequence $s_1$, $s_2$, ..., $s_k$ going up to the root, because $NL'_k$ is formed by (2) with the following two N-lists:

- $NL_{k-1}$ with property (*) at step 2
- $NL'$ contains all N-nodes associated with selector $s_k$.

Assume now that the subset relation in (**) is proper, i.e. that there exists an N-node in $NL'_k$ that is not in $NL_k$. In other words, the sum of the frequency counts of all nodes in $NL_k$ is less than that of $NL'_k$. However, (***) implies that the sum of the frequency counts of all nodes in $NL'_k$ is the support count of the $k$-selector-set $s_1 s_2 \cdots s_k$. This infers that Property 3 does not hold for N-lists generated with Definition 7(i), which is a contradiction. Therefore, $NL_k = NL'_k$, and the lemma holds. □

Note that this is not a trivial result. In fact, Property 3 does not necessarily hold for all possible ways of joining two N-lists. Some counter-examples can be found in Fig. 1:

***Example 1*** $s_{12}s_{22} \rightarrow \{\langle 2, 4\rangle : 1, \langle 14, 17\rangle : 2\}$ $s_{32}s_2 \rightarrow \{\langle 13, 18\rangle : 2\}$ The N-list of selector-set $s_{12}s_{32}s_{22}s_2$ generated from the above N-lists of $s_{12}s_{22}$ and $s_{32}s_2$ using equation (2) of Definition 7 is $\{\langle 13, 18\rangle : 2\}$. This does not correctly capture the number of paths that contain all four selectors, because while two paths go through $s_{12}s_{22}s_2$ in the right-hand branch of the PC-tree, only the center path also goes through $s_{32}$. The correct N-list would thus be $\{\langle 13, 18\rangle : 1\}$.

***Example 2*** $s_{12}s_{22} \rightarrow \{\langle 2, 4\rangle : 1, \langle 14, 17\rangle : 2\}$ $s_{11} \rightarrow \{\langle 5, 3\rangle : 1; \langle 8, 10\rangle : 3; \langle 18, 16\rangle : 1\}$ The N-list of selector-set $s_{12}s_{11}s_{22}$ generated from the above N-lists of $s_{11}$ and $s_{12}s_{22}$ using equation (2) of Definition 7 is $\{\langle 2, 4\rangle : 1; \langle 14, 17\rangle : 1\}$ which does not correctly capture the number of paths containing all three selectors. The reason is that two of the five paths that reach $s_{11}$ in the PPC-tree go through $s_{22}$, but neither of them continues through $s_{12}$. The correct N-list would thus be { }.

**Acknowledgements** The authors are very grateful to the editor and three very thorough reviewers, whose detailed comments and suggestions substantially helped to improve this paper.

**Author contributions** Van Quoc Phuong Huynh conceived the basic idea behind the proposed algorithm and implemented it. Johannes Fürnkranz placed the algorithm in the context of prior work in rule learning. All authors contributed to the design of the experiments, the interpretation of the results and the write-up of the paper.

**Funding** Open access funding provided by Johannes Kepler University Linz.

**Data availability** The datasets we used can be found at https://archive.ics.uci.edu/ml/index.php.

**Code availability** The code of the Lord algorithm can be found at https://github.com/vqphuynh/LORD.

**Declarations**

**Conflict of Interest** All auhtor declares no conflict of Interest.